\documentclass{article}

\usepackage[final]{neurips_2022}

\usepackage[utf8]{inputenc} 
\usepackage[T1]{fontenc}    
\usepackage[colorlinks=true,%
linkcolor= blue,
anchorcolor=blue,%
citecolor=blue,
filecolor=blue,%
menucolor=blue,%
urlcolor=blue]{hyperref}
\usepackage{url}            
\usepackage{booktabs}       
\usepackage{amsfonts}       
\usepackage{nicefrac}       
\usepackage{microtype}      
\usepackage{xcolor}         

\usepackage{tikz}
\usepackage[ruled,vlined,linesnumbered]{algorithm2e}
\usepackage{caption}
\usepackage{subcaption}

 \PassOptionsToPackage{round}{natbib}

\usepackage{booktabs}

\usepackage{color}
\usepackage{latexsym}              
\usepackage{amsmath}               
\usepackage{amssymb}               
\usepackage{amsthm}
\usepackage{multirow}
\usepackage{tikz}                  
\usetikzlibrary{arrows,positioning,shapes}
\usepackage{tcolorbox}

\usepackage[mathcal]{eucal}
\usepackage{cleveref}
\crefname{equation}{Eq.}{Eqs.}
\crefname{table}{Table}{Tables}
\crefname{section}{Sec.}{Secs.}
\crefname{theorem}{Thm.}{Thms.}
\crefname{lemma}{Lemma}{Lemmas}
\crefname{corollary}{Corollary}{Cors.}
\crefname{appendix}{Appendix}{Appendixes}
\crefname{remark}{Remark}{Remark}
\crefname{figure}{Fig.}{Figs.}

\newtheorem{theorem}{Theorem}[section]
\newtheorem{definition}[theorem]{Definition}
\newtheorem{lemma}[theorem]{Lemma}
\newtheorem{proposition}[theorem]{Proposition}
\newtheorem{corollary}[theorem]{Corollary}
\newtheorem{remark}[theorem]{Remark}

\newcommand{\OPBTest}{\normalfont\textsc{OPBTest} }
\newcommand{\OPBTrain}{\normalfont\textsc{OPBTrain} }

\usepackage{dsfont}

\PassOptionsToPackage{round}{natbib}

\title{Online PAC-Bayes Learning}

\author{%
  Maxime Haddouche \\
  Inria and University College London\\
  France and UK\\
   \And
   Benjamin Guedj \\
  Inria and University College London \\
   France and UK \\
}

\begin{document}

\date{}

\maketitle

\begin{abstract}
   Most PAC-Bayesian bounds hold in the batch learning setting where data is collected at once, prior to inference or prediction. This somewhat departs from many contemporary learning problems where data streams are collected and the algorithms must dynamically adjust. We prove new PAC-Bayesian bounds in this online learning framework, leveraging an updated definition of regret, and we revisit classical PAC-Bayesian results with a batch-to-online conversion, extending their remit to the case of dependent data. Our results hold for bounded losses, potentially \emph{non-convex}, paving the way to promising developments in online learning.
\end{abstract}

\section{Introduction}

 Batch learning is somewhat the dominant learning paradigm in which we aim to design the best predictor by collecting a training dataset which is then used for inference or prediction. Classical algorithms such as SVMs \citep[see][among many others]{cristianini2000introduction} or feedforward neural networks \citep{svozil1997introduction} are popular examples of efficient batch learning. While the mathematics of batch learning constitute a vivid and well understood research field, in practice this might not be aligned with the way practitionners collect data, which can be sequential when too much information is available at a given time (\emph{e.g.} the number of micro-transactions made in finance on a daily basis). Indeed batch learning is not designed to properly handle dynamic systems.

 Online learning (OL)  \citep{zinkevich2003online,shalev2012online,hazan2016introduction} fills this gap by treating data as a continuous stream with a potentially changing learning goal.
 OL has been studied with convex optimisation tools and the celebrated notion of regret which measures the discrepancy between the cumulative sum of losses for a specific algorithm at each datum and the optimal strategy. It led to many fruitful results comparing the  efficiency of prediction for optimisation algorithms such that Online Gradient Descent (OGD), Online Newton Step through static regret \citep{zinkevich2003online,hazan2007logarithmic}. OL is flexible enough to incorporate external expert advice onto classical algorithms with the optimistic point of view that such advices are useful for training \citep{rakhlin2013online,rakhlin2013practical} and then having optimistic regret bounds.
 Modern extensions also allow to compare to moving strategies through dynamic regret \citep[see e.g.][]{yang2016tracking,zhang2018strong,zhao2020dyn}. However, this notion of regret has been challenged recently: for instance, \citet{wintenberger2021stochastic} chose to control an expected cumulative loss through PAC inequalities in order to deal with the case of stochastic loss functions.

Statements holding with arbitrarily large probability are widely used in learning and especially within the PAC-Bayes theory. Since its emergence in the late 90s, the PAC-Bayes theory (see the seminal
works of \citealp{STW1997}, \citealp{McAllester1998, McAllester1999} and the recent surveys by \citealp{guedj2019primer,alquier2021survey}) has been a powerful tool to obtain generalisation bounds and to derive efficient learning algorithms. Classical PAC-Bayes generalisation bounds help to understand how a learning algorithm may perform on future similar batches of data.
More precisely, PAC-Bayes learning exploits the Bayesian paradigm of explaining a learning problem through a meaningful distribution over a space of candidate predictors
\citep[see e.g.][]{maurer2004note,catoni2007pac,seldin2013pac,mhammedi2019}.
An active line of research in PAC-Bayes learning is to overcome classical assumptions such that data-free prior, bounded loss, iid data \citep[see][]{lever2010,lever2013,Alquier_2017,holland2019,rivasplata2020pac,haddouche2021pac,pujol2021pac} while remaining in a batch learning sprit.
Finally, a pioneering line of work led by \citep{seldin2012bandit,seldin2012marting} on PAC-Bayes learning for martingales and independently developed by \citep{gerchinovitz2011sparse,foster2015adaptive,li2018pacbayes} boosted PAC-Bayes learning by providing sparsity regret bound, adaptive regret bounds and online algorithms for clustering.

\textbf{Our contributions.} Our goal is to provide a general online framework for PAC-Bayesian learning. Our main contribution (\cref{th: main_thm_online} in \cref{sec: main_bound}) is a general bound  which is then used to derive several online PAC-Bayesian results (as developed in \cref{sec: online_pacb_procedure,sec: OPBD_procedure}).
More specifically, we derive two types of bounds, \emph{online PAC-Bayesian training and test bounds}.
Training bounds exhibit online procedures while the test bound provide efficiency guarantees. We propose then several algorithms with their associated training and test bounds as well as a short series of experiments to evaluate the consistency of our online PAC-Bayesian approach. Our efficiency criterion is not the classical regret but an expected cumulative loss close to the one of \citet{wintenberger2021stochastic}. More precisely, \cref{sec: online_pacb_procedure} propose a stable yet time-consuming Gibbs-based algorithm, while \cref{sec: OPBD_procedure} proposes time efficient yet volatile algorithms.
We emphasize that our PAC-Bayesian results only require a bounded loss to hold: no assumption is made on the data distribution, priors can be data-dependent and we do not require any convexity assumption on the loss, as commonly assumed in the OL framework.

\textbf{Outline.} \cref{sec: main_bound} introduces the  theoretical framework as well as our main result. \cref{sec: online_pacb_procedure} presents an online PAC-Bayesian algorithm and draws links between PAC-Bayes and OL results.
\cref{sec: OPBD_procedure} details online PAC-Bayesian disintegrated procedures with reduced computational time and \cref{sec: experiments} gathers supporting experiments.
We include reminders on OL and PAC-Bayes in \cref{sec: OGD_reminder,sec: disintegrated_bounds}. \cref{sec: discussion_main_thm} provide disucssion about our main result. All proofs are deferred to \cref{sec: proofs}.

\section{An online PAC-Bayesian bound}
\label{sec: main_bound}
We establish a novel PAC-Bayesian theorem (which in turn will be particularised in \cref{sec: online_pacb_procedure}) which overcomes the classical limitation of data-independent prior and iid data.
We call our main result an \emph{online PAC-Bayesian bound} as it allows to consider a sequence of priors which may depend on the past and a sequence of posteriors that can dynamically evolve as well. Indeed, we follow the online learning paradigm which considers a continous stream of data that the algorithm has to process on the fly, adjusting its outputs at each time step w.r.t. the arrival of new data and the past. In the PAC-Bayesian framework, this paradigm translates as follows: from an initial (still data independent) prior $Q_1=P$ and a data sample $S= (z_1,...,z_m)$, we design a sequence of posterior $(Q_i)_{1\leq i\leq m }$ where $Q_i= f(Q_1,...,Q_{i-1},z_i)$.

\textbf{Framework.} Consider a data space $\mathcal{Z}$ (which can be only inputs or pairs of inputs/outputs). We fix an integer $m>0$ and our data sample $S\in\mathcal{Z}^m$ is drawn from an unknown distribution $\mu$. We do not make any assumption on $\mu$.  We set a sequence of priors, starting with $P_1=P$ a data-free distribution and $(P_i)_{i\geq 2}$ such that for each $i$, $P_i$ is $\mathcal{F}_{i-1}$ measurable where $(\mathcal{F}_i)_{i\geq 0}$ is an adapted filtration to $S$.
For $P,Q \in \mathcal{M}_{1}\left(\mathcal{H}\right)$, the notation $P \ll Q$ indicates that $Q$ is absolutely continuous wrt $P$ (i.e. $Q(A) = 0$ if $P(A) = 0$ for measurable $A \subset \mathcal{H}$).
We also denote by $Q_i$ our sequence of candidate posteriors. There is no restriction on what $Q_i$ could be.
In what follows we fix a filtration $(\mathcal{F}_i)_{i\geq 0}$ and we denote by $\operatorname{KL}$ the Kullback-Leibler divergence between two distributions.

We consider a predictor space $\mathcal{H}$ and a loss funtion $\ell: \mathcal{H}\times \mathcal{Z} \rightarrow \mathbb{R}^+$ bounded by a real constant $K>0$. This loss defines the (potentially moving) learning objective. We denote by $\mathcal{M}_1(\mathcal{H})$ the set of all probability distributions on $\mathcal{H}$.
 We now introduce the notion of \emph{stochastic kernel} \citep{rivasplata2020pac} which formalise properly data-dependent measures within the PAC-Bayes framework. First, for a fixed predictor space $\mathcal{H}$, we set $\Sigma_{\mathcal{H}}$ to be the considered $\sigma$-algebra on $\mathcal{H}$.

\begin{definition}[Stochastic kernels]
    A \emph{stochastic kernel} from $\mathcal{S}=\mathcal{Z}^m$ to $\mathcal{H}$ is defined as a mapping $Q: \mathcal{Z}^m\times \Sigma_{\mathcal{H}} \rightarrow [0;1]$ where
    \begin{itemize}
        \item For any $B\in \Sigma_{\mathcal{H}}$, the function  $S=(z_1,...,z_m)\mapsto Q(S,B)$ is measurable,
        \item For any $S\in\mathcal{Z}^m$, the function $B\mapsto Q(S,B)$ is a probability measure over $\mathcal{H}$.
    \end{itemize}
    We denote by $\texttt{Stoch}(\mathcal{S},\mathcal{H})$ the set of all stochastic kernels from $\mathcal{S}$ to $\mathcal{H}$ and for a fixed $S$, we set $Q_S:= Q(S,.)$ the data-dependent prior associated to the sample $S$ through Q.
\end{definition}

From now, to refer to a distribution $Q_S$ depending on a dataset $S$, we introduce a stochastic kernel $Q(.,.)$ such that $Q_S = Q(S,.)$. Note that this notation is perfectly suited to the case when $Q_S$ is obtained from an algorithmic procedure $A$. In this case the stochastic kernel $Q$ of interest is the learning algorithm $A$.
We use this notion to characterise our sequence of priors.

\begin{definition}
  We say that a sequence of stochastic kernels $(P_i)_{i=1..m}$ is an \emph{\textbf{online predictive sequence}} if (i) for all $i\geq 1, S\in\mathcal{Z}^m, P_i(S,.)$ is $\mathcal{F}_{i-1}$ measurable and (ii) for all $i \geq 2$, $P_i(S,.)\gg P_{i-1}(S,.)$.
\end{definition}

Note that (ii) implies that for all $i, P_i(S,.)\gg P_1(S,.)$ with $P_1(S,.)$ a data-free measure (yet a classical prior in the PAC-Bayesian theory).

We can now state our main result.

\begin{theorem}
  \label{th: main_thm_online}
  For any distribution $\mu$ over $\mathcal{Z}^m$, any $\lambda>0$ and any online predictive sequence (used as priors) $(P_i)$, for any sequence of stochastic kernels $(Q_i)$ we have with probability $1-\delta$ over the sample $S\sim\mu$, the following, holding for the data-dependent measures $Q_{i,S}:= Q_i(S,.), P_{i,S}:= P_i(S,.)$ :

  \[ \sum_{i=1}^m \mathbb{E}_{h_i\sim Q_{i,S}}\left[ \mathbb{E}[\ell(h_i,z_i) \mid \mathcal{F}_{i-1}]    \right]  \leq \sum_{i=1}^m \mathbb{E}_{h_i\sim Q_{i,S}}\left[ \ell(h_i,z_i) \right] + \frac{\operatorname{KL}(Q_{i,S}\| P_{i,S})}{\lambda} + \frac{\lambda m K^2}{2} + \frac{\log(1/\delta)}{\lambda}. \]

\end{theorem}

\begin{remark}
  \label{rem: notations}
  For the sake of clarity, we assimilate in what follows the stochastic kernels $Q_i,P_i$ to the data-dependent distributions $Q_i(S,.), P_i(S,.)$. Then, an online predictive sequence is also assimilated to a sequence of data-dependent distributions. Concretely this leads to the switch of notation $Q_{i,S}\rightarrow Q_i$ in \cref{th: main_thm_online}. The reason of this switch is that, even though stochastic kernel is the right theoretical structure to state our main result, we consider in \cref{sec: online_pacb_procedure,sec: OPBD_procedure} practical algorithmic extensions which focus only on data-dependent distributions, hence the need to alleviate our notations.
\end{remark}

The proof is deferred to \cref{sec: proof_main_thm_online}. See \cref{sec: discussion_main_thm} for context and discussions.

\textbf{A batch to online conversion.}
   First, we remark that our bound slightly exceeds the OL framework: indeed, it would require our posterior sequence to be an online predictive sequence as well, which is not the case here (for any $i$, the distribution $Q_{i,S}$ can depend on the whole dataset ). This is a consequence of our proof method (see \cref{sec: proof_main_thm_online}), which is classically denoted as a "batch to online" conversion (in opposition to the "online to batch" procedures as in \citealp{dekel2005data}). In other words, we exploited PAC-Bayesian tools designed for a fixed batch of data to obtain a dynamic result. This is why we refer to our bound as online as it allows to consider sequences of priors and posteriors that can dynamically evolve.

\textbf{Analysis of the different terms in the bound.}
Our PAC-Bayesian bound formally differs in many points from the classical ones.
  On the left-hand side of the bound, the sum of the averaged expected loss conditioned to the past appears. Having such a sum of expectations instead of a single one is necessary to assess the quality of all our predictions. Indeed, because data may be dependent, one can not consider a single expectation as in the iid case. We also stress that taking an online predictive sequence as priors leads to control losses conditioned to the past, which differs from classical PAC-Bayes results designed to bound the expected loss. This term, while original in the PAC-Bayesian framework (to the best of our knowledge) recently appeared (in a modified form) in \citet[Prop 3]{wintenberger2021stochastic}. See \cref{sec: deeper_analysis_main_thm} for further disucssions.

  On the right hand-side of the bound, online counterparts of classical PAC-Bayes terms appear. At time $i$, the measure $Q_i$ (i.e. $Q_{i,S}$ according to \cref{rem: notations}) has a tradeoff to achieve between an overfitted prediction of $z_i$ (the case $Q_i=\delta_{z_i}$ where $\delta$ is a Dirac measure) and a too weak impact of the new data with regards to our prior knowledge (the case $Q_i=P_i$). The quantity $\lambda>0$ can be seen as a regulariser to adjust the relative impact of both terms.

\textbf{Influence of $\lambda$.}
The quantity $\lambda$ also plays a crucial role on the bound as it is involved in an explicit tradeoff between the KL terms, the confidence term $\log(1/\delta)$ and the residual term $mK^2/2$. This idea of seeing $\lambda$ as a trading parameter is not new \citep{thiemann2017strongly,germain2016pac}.
 However, the results from \citet{thiemann2017strongly} stand w.p. $1-\delta $ for any $\lambda$ while ours and the ones from \citet{germain2016pac} hold for any $\lambda$ w.p. $1-\delta$ which is weaker and implies to discretise $\mathbb{R}^+$ onto a grid to estimate the optimal $\lambda$.

We now move on to the design of online PAC-Bayesian algorithms.

\section{An online PAC-Bayesian procedure}

\label{sec: online_pacb_procedure}

OL algorithms (we refer to \citealp{hazan2016introduction}  an introduction to the field) are producing sequences of predictors by progressively updating the considered predictor (see \cref{sec: OGD_reminder} for an example). Recall that, in the OL framework, an algorithm outputs at time $i$ a predictor which is $\mathcal{F}_{i-1}$-measurable. Here, our goal is to design an online procedure derived from \cref{th: main_thm_online} which outputs an online predictive sequence (which is assimilated, according to \cref{rem: notations}, to a sequence of distributions).

\textbf{Online PAC-Bayesian (OPB) training bound.} We state a corollary of our main result which paves the way to an online algorithm. This constructive procedure motivates the name \emph{ Online PAC-Bayesian training bound} (\OPBTrain in short).

\begin{corollary}[\normalfont\textsc{OPBTrain}]
  \label{cor: online_procedure}
  For any distribution $\mu$ over $\mathcal{Z}^m$, any $\lambda>0$ and any online predictive sequences $\hat{Q},P$, the following holds with probability $1-\delta$ over the sample $S\sim \mu$ :

  \begin{align*}
    \sum_{i=1}^m \mathbb{E}_{h_i\sim \hat{Q}_{i+1}}\left[ \mathbb{E}[\ell(h_i,z_i) \mid \mathcal{F}_{i-1}]    \right] \leq \sum_{i=1}^m \mathbb{E}_{h_i\sim \hat{Q}_{i+1}}\left[ \ell(h_i,z_i) \right] + \frac{\operatorname{KL}(\hat{Q}_{i+1}\| P_i)}{\lambda} + \frac{\lambda m K^2}{2} + \frac{\log(1/\delta)}{\lambda}.
  \end{align*}
\end{corollary}
Here, $\lambda$ is seen as a scale parameter as precised below.
The proof consists in applying \cref{th: main_thm_online} with for all $i$, $Q_i= \hat{Q}_{i+1}$ and $P_i$.
Note that in this case, our posterior sequence is an online predictive sequence in order to fit with the OL framework.

\cref{cor: online_procedure} suggests to design $\hat{Q}$ as follows, assuming we have drawn a dataset $S= \{z_1,...,z_m\}$, fixed a scale parameter $\lambda>0$  and an online predictive sequence $P_i$:

\begin{align}
  \label{eq: pacb_online_alg}
  \hat{Q}_1= P_1, \quad \forall i\geq1\; \hat{Q}_{i+1}&= \underset{Q\in\mathcal{M}_1(\mathcal{H})}{\mathrm{argmin}} \mathbb{E}_{h_i\sim Q} \; [\ell(h_i,z_i)] + \frac{\operatorname{KL}(Q\| P_i)}{\lambda} \\
  \intertext{which leads to the explicit formulation}
  \label{eq: catoni_formulation}
  \frac{d\hat{Q}_{i+1}}{dP_i}(h)& = \frac{\exp\left(-\lambda \ell(h,z_i)\right)}{\mathbb{E}_{h\sim P_i}\left[\exp\left(-\lambda \ell(h,z_i)\right)\right]}.
\end{align}

  Thus, the formulation of \cref{eq: catoni_formulation}, which has been highlighted by \citet[Sec. 5.1]{catoni2003pac} shows that our online procedure produces Gibbs posteriors.
 So, PAC-Bayesian theory provides sound justification for the somewhat intuitive online procedure in \cref{eq: pacb_online_alg}: at time $i$, we adjust our new measure $\hat{Q}_{i+1}$ by optimising a tradeoff between the impact of the newly arrived data $z_i$ and the one of prior knowledge $\hat{Q}_i$.

 Notice that $\hat{Q}$ is an online predictive sequence: $\hat{Q}_i$ is $\mathcal{F}_{i-1}$-measurable for all $i$ as it depends only on $\hat{Q}_{i-1}$ and $z_{i-1}$. Furthermore, one has $\hat{Q}_{i} \gg \hat{Q}_{i-1}$ for all $i$ as $\hat{Q}_{i}$ is defined as an argmin and the KL term is finite if and only it is absolutely continuous w.r.t. $\hat{Q}_{i-1}$.

\begin{remark}
  In \cref{cor: online_procedure}, while the right hand-side is the reason we considered \cref{eq: pacb_online_alg}, the left hand side still needs to be analysed. It expresses how the posterior $\hat{Q}_{i+1}$ (designed from $\hat{Q}_i,z_i$) generalises well on average to any new draw of $z_i$. More precisely, this term measures how much the training of $\hat{Q}_{i+1}$ is overfitting on $z_i$. A low value of it ensures our online predictive sequence, which is obtained from a single dataset, is robust to the randomness of $S$, hence the interest of optimising the right hand side of the bound.
  This is a supplementary reason we refer to \cref{cor: online_procedure} as an \OPBTrain bound as it provide robustness guarantees for our training.
\end{remark}

\textbf{Online PAC-Bayesian (OPB) test bound.}
However, \cref{cor: online_procedure} does not say if $\hat{Q}_{i+1}$ will produce good predictors to minimise $\ell(.,z_{i+1})$, which is the objective of $\hat{Q}_{i+1}$ in the OL framework (we only have access to the past to predict the future). We then need to provide an \emph{Online PAC-Bayesian (OPB) test bound} (\OPBTest bound) to quantify our prediction's accuracy. We now derive an \OPBTest bound from \cref{th: main_thm_online}.

\begin{corollary}[\normalfont\textsc{OPBTest}]
  \label{cor: test_bound_online}.
  For any distribution $\mu$ over $\mathcal{Z}^m$, any $\lambda>0$, and any online predictive sequence $(\hat{Q}_i)$, the following holds with probability $1-\delta$ over the sample $S\sim \mu$:
  \begin{align*}
    \sum_{i=1}^m \mathbb{E}_{h_i\sim \hat{Q}_{i}}\left[ \mathbb{E}[\ell(h_i,z_i) \mid \mathcal{F}_{i-1}]    \right] \leq \sum_{i=1}^m \mathbb{E}_{h_i\sim \hat{Q}_{i}}\left[ \ell(h_i,z_i) \right] + \frac{\lambda m K^2}{2} + \frac{\log(1/\delta)}{\lambda}.
  \end{align*}
  Optimising in $\lambda$ gives $\lambda= \sqrt{\nicefrac{2\log(1/\delta)}{mK^2} }$ and ensure that:
  \begin{align*}
    \sum_{i=1}^m \mathbb{E}_{h_i\sim \hat{Q}_{i}}\left[ \mathbb{E}[\ell(h_i,z_i) \mid \mathcal{F}_{i-1}]    \right] \leq \sum_{i=1}^m \mathbb{E}_{h_i\sim \hat{Q}_{i}}\left[ \ell(h_i,z_i) \right] + O\left(\sqrt{\log(1/\delta) K^2m}\right).
  \end{align*}

\end{corollary}

The proof consists in applying \cref{th: main_thm_online} with for all $i$, $Q_i= \hat{Q}_{i}= P_i$.

\cref{cor: test_bound_online} quantifies how efficient will our predictions be. Indeed, the left hand side of this bound relates for all $i$, how good $\hat{Q}_i$ is to predict $z_i$ (on average) which is what $\hat{Q}_i$ is designed for.
Note that here, the involved $\lambda$ can differ from the scale parameter of \cref{eq: pacb_online_alg}, it is now a way to compensate for the tradeoff between the two last terms of the bound. The strength of this bound is that since $\hat{Q}$ is an online predictive sequence, the Kullback-Leibler terms vanished, leaving terms depending only on hyperparameters.

\subsection*{Links with previous approaches}

We now present a specific case of \cref{cor: online_procedure} where we choose as priors the online predictive sequence $\hat{Q}$ (\emph{i.e.} in \cref{th: main_thm_online}, we choose $Q_i=\hat{Q}_{i+1}, P_i= \hat{Q}_i$). The reason we focus on this specific case is that it enables to build strong links between PAC-Bayes and OL.

We then adapt our \OPBTrain bound (\cref{cor: online_procedure}). The online procedure becomes:
\begin{align}
  \label{eq: pacb_online_alg_specific_case}
  \hat{Q}_1= P, \quad \forall i\geq1\; \hat{Q}_{i+1}&= \operatorname{argmin}_{Q} \mathbb{E}_{h_i\sim Q} \; [\ell(h_i,z_i)] + \frac{\operatorname{KL}(Q\| \hat{Q}_{i})}{\lambda},
\end{align}
which leads to the explicit formulation
\begin{align*}
  \frac{d\hat{Q}_{i+1}}{d\hat{Q}_{i}}(h) = \frac{\exp\left(-\lambda \ell(h,z_i)\right)}{\mathbb{E}_{h\sim \hat{Q}_{i}}\left[\exp\left(-\lambda \ell(h,z_i)\right)\right]}.
\end{align*}

\textbf{Links with classical PAC-Bayesian bounds.} We denote that the optimal predictor in this case is such that at any time $i$, $d\hat{Q}_{i+1}(h) \propto \exp(-\lambda\ell(h,z_i))d\hat{Q}_i(h) $ hence $d\hat{Q}_{m+1}(h) \propto \exp\left(-\lambda\sum_{i=1}^m\ell(h,z_i)\right)d\hat{Q}_1(h) $.
One recognises, up to a multiplicative constant, the optimised predictor of \citet[][Th 1.2.6]{catoni2007pac} which solves
$\operatorname{argmin}_{Q} \mathbb{E}_{h\sim Q} \; [\frac{1}{m}\sum_{i=1}^m\ell(h,z_i)] + \frac{\operatorname{KL}(Q\| \hat{Q}_{1})}{\lambda}$, thus one sees that in this case, the output of our online procedure after $m$ steps coincides with Catoni's output. This shows consistency of our general procedure which recovers classical result within an online framework: when too many data are available, treating data sequentially until time $m$ leads to the same Gibbs posterior than if we were treating the whole dataset as a batch.

\textbf{Analogy with Online Gradient Descent (OGD).} We propose an analogy between the procedure \cref{eq: pacb_online_alg_specific_case} and the celebrated OGD algorithm (see \cref{sec: OGD_reminder} for a recap). First we remark that our minimisation problem is equivalent to
$\operatorname{argmin}_{Q} \lambda\mathbb{E}_{h_i\sim Q} \; [\ell(h_i,z_i)] + \operatorname{KL}(Q\| \hat{Q}_{i})$.
Then we assume that for any $i, \hat{Q}_i=\mathcal{N}(\hat{m}_i,I_d)$ with $\hat{m_i}\in\mathbb{R}^d$  and we set $\mathcal{L}_i(\hat{m}_i)= \mathbb{E}_{h_i\sim \hat{Q}_i} \; [\ell(h_i,z_i)] $ .
The minimisation problem becomes: $\operatorname{argmin}_{\hat{m}} \lambda\mathcal{L}_i(\hat{m}) + \frac{1}{2} \| \hat{m} - \hat{m}_i \|^2$.
And so using the first order Taylor expansion, we use the approximation $ \mathcal{L}_i(\hat{m}) \approx \mathcal{L}_i(\hat{m}_i ) + \langle \hat{m}- \hat{m_i}, \nabla \mathcal{L}_i(\hat{m}_i) \rangle $ which finally transform our argmin into the following optimisation process: $\hat{m}_{i+1} = \hat{m_i} - \lambda \nabla \mathcal{L}_i(\hat{m}_i)$ which is exactly OGD on the loss sequence $\mathcal{L}_i$.
We draw an analogy between the scale parameter $\lambda$ and the step size $\eta$ in OGD. the KL term translates the influence of the previous point and the expected loss gives the gradient.
This analogy has been already exploited in \citet{shalev2012online} where they approximated $\mathbb{E}_{h_i\sim q_\mu} [\ell(h_i,z_i)]:= \bar{L}_i(\mu) \approx \mu^T\nabla \bar{L}_i(\mu_i)$ where $\mu$ is their considered online predictive sequence.

Finally, we remark that the optimum rate in \cref{cor: test_bound_online} is a $O(\sqrt{m})$ which is comparable to the best rate of \citet[][Eq (2.5)]{shalev2012online} (see \cref{prop: OGD_bound}).

\textbf{Comparison with previous work.} We acknowledge that the procedure of \cref{eq: pacb_online_alg_specific_case} already appeared in literature. \citet[][Alg. 1]{li2018pacbayes} propose a Gibbs procedure somewhat similar to ours, the main difference being the addition of a surrogate of the true loss at each time step.
Within the OL literature, the idea of updating measures online has been recently studied for instance in \citet{cherief2019generalization}. More precisely, our procedure is similar to their Streaming Variational Bayes (SVB) algorithm. A slight difference is that they approximated the expected loss similarly to \citet{shalev2012online}.
The guarantees \citet{cherief2019generalization} provided for SVB hold for Gaussian priors and comes at the cost of additional constraints that do not allow to consider any aggregation strategies contrary to what \cref{cor: online_procedure} propose. Their bounds are deterministic and are using tools and assumptions from convex optimisation (such that convex expected losses) while ours are probabilistic and are using measure theory tools which allow to relax these assumptions.

\textbf{Strength of our result.} We emphasize two points. First, to the best of our knowledge, \cref{cor: online_procedure} is the first bound which theoretically suggests \cref{eq: pacb_online_alg_specific_case} as a learning algorithm.
Second, we stress that \cref{eq: pacb_online_alg_specific_case}
is a particular case of \cref{cor: online_procedure} and our result can lead to other fruitful routes. For instance, we consider the idea of adding noise to our measures at each time step to avoid overfitting (this idea has been used \emph{e.g.} in \citealp{neelakantan2015adding} in the context of deep neural networks): if our online predicitve sequence $(\hat{Q}_i)$ can be defined through a sequence of parameter vectors $\hat{\mu}$, then we can define $P_i$ by adding a small noise on $\hat{\mu}_i$ and thus giving more freedom through stochasticity.

Thus, we see that our procedure led us to the use of the Gibbs posteriors of Catoni. However, in practice, Gaussian distributions are preferred \citep[\emph{e.g.}][]{dziugaite2017computing, rivasplata2019pac,perezortiz2021progress,perezortiz2021learning,perez2021tighter}).
That is why we focus next on new online PAC-Bayesian algorithms involving Gaussian distributions.

\section{Disintegrated online algorithms for Gaussian distributions.}
\label{sec: OPBD_procedure}
We dig deeper in the field of disintegrated PAC-Bayesian bounds, originally explored by \citet{catoni2007pac,blanchard2007occam}, further studied by \citet{alquier2013sparse,guedj2013} and recently developed by \citet{rivasplata2020pac,viallard2021general} (see \cref{sec: disintegrated_bounds} for a short presentation of the bound we adapted and used). The strength of the disintegrated approach is that we have directly guarantees on the random draw of a single predictor, which avoids to consider expectations over the predictor space.
This fact is particularly significant in our work as the procedure precised in \cref{eq: catoni_formulation}, require the estimation of an exponential moment to be efficient, which may be costful.
We then show that disintegrated PAC-Bayesian bounds can be adapted to the OL framework, and that they have the potential to generate proper online algorithms with weak computational cost and sound efficiency guarantees.

\textbf{Online PAC-Bayesian disintegrated (OPBD) training bounds.} We present a general form for \emph{online PAC-Bayes disintegrated (OPBD) training bounds}. The terminology comes from the way we craft those bounds: from PAC-Bayesian disintegrated bounds we use the same tools as in \cref{th: main_thm_online} to create the first online PAC-Bayesian disintegrated bounds.
OPBD training bounds have the following form.

For any online predictive sequences $\hat{Q},P$, any $\lambda>0$ w.p. $1-\delta$ over $S\sim \mu$ and $(h_1,...,h_{m})\sim \hat{Q}_2\otimes...\otimes \hat{Q}_{m+1}$:
\begin{align}
  \label{eq: OPBD_train_bound}
  \sum_{i=1}^m  \mathbb{E}[\ell(h_i,z_i) \mid \mathcal{F}_{i-1}]   \leq \sum_{i=1}^m  \ell(h_i,z_i)  + \Psi(h_{i},\hat{Q}_{i+1},P_i) \; + \Phi(m),
\end{align}

with $\Psi,\Phi$ being real-valued functions. $\Psi$ controls the global behaviour of $Q_{i+1}$ w.r.t. the $\mathcal{F}_{i-1}$-measurable prior $P_{i}$. If one has no dependency on $h_i$ this behaviour is global, otherwise it is local.
Note that those functions may depend on $\lambda,\delta$. However, since they are fixed parameters, we do not make these dependencies explicit.
Similarly to \cref{cor: online_procedure}, this kind of bound allows to derive a learning algorithm (cf \Cref{alg: OPBD_alg}) which outputs an online predicitve sequence $\hat{Q}$.
Finally we draw $(h_1,...,h_{m})\sim \hat{Q}_2\otimes...\otimes \hat{Q}_{m+1}$ (and not $\hat{Q}_1\otimes...\otimes \hat{Q}_{m}$)  since an OPBD bound is designed to justify theoretically an OPBD procedure in the same way \cref{cor: online_procedure} allowed to justify \cref{eq: pacb_online_alg}.

\textbf{Why focus on Gaussian measures?} The reason is that a Gaussian variable $h\sim\mathcal{N}(w,\sigma^2\mathbf{I}_d)$ can be written as $h=w +\varepsilon$ with $\varepsilon\sim\mathcal{N}(0,\sigma^2\mathbf{I}_d)$, and this expression totally defines $h$ ($\mathbf{I}_d$ being the identity matrix).

\textbf{A general OPBD algorithm for Gaussian measure with fixed variance} We use an idea presented in \citet{viallard2021general} which restrict the measure set to Gaussian on $\mathbb{R}^d$ \textbf{with known and fixed covariance matrix} $\sigma^2 \mathbf{I}_d$.
Then we present in \Cref{alg: OPBD_alg} a general algorithm (derived from an OPBD training bound) for Gaussian measures with fixed variance which outputs a sequence of gaussian $\hat{Q}_i=\mathcal{N}(\hat{w}_i,\sigma^2\mathbf{I}_d)$ from a prior sequence $P_i= \mathcal{N}(w_i^0,\sigma^2\mathbf{I}_d)$
where for each $i$, $w_i^0$ is $\mathcal{F}_{i-1}$- measurable. Because the variance is fixed, the distribution is uniquely defined by its mean, thus we identify $\hat{Q}_i$ and $\hat{w}_i$, $P_i$ and $w_i^0$.

\begin{algorithm}[ht]
 \SetAlgoLined
 \SetKwInOut{Initialisation}{Initialisation}
 \SetKwInOut{Parameter}{Parameters}
 \Parameter{Time m, scale parameter $\lambda$ }
 \Initialisation{Variance $\sigma^2$, Initial mean $\hat{w}_1\in\mathbb{R}^d$, epoch $m$}
\For{ each iteration $i$ in $1..m$}{ Observe $z_i, w_i^0$ and draw $\varepsilon_i\sim \mathcal{N}(0,\sigma^2\mathbf{I}_d)$\\
Update:
\[ \hat{w}_{i+1}:= \operatorname{argmin}_{w\in\mathbb{R}^d} \ell(w + \varepsilon_i,z_i) + \Psi(w+ \varepsilon_i,w, w_i^0)   \]}
\textbf{Return} $(\hat{w}_i)_{i=1..m+1}$
 \caption{A general OPBD algorithm for Gaussian measures with fixed variance.}
 \label{alg: OPBD_alg}
 \end{algorithm}

At each time $i$, \Cref{alg: OPBD_alg} requires the draw of $\varepsilon_i\sim\mathcal{N}(0,\sigma^2 \mathbf{I}_d)$. Doing so, we generated the randomness for our $h_i$ (because our bound holds for a single draw of $(h_1,..,h_m)\sim \hat{Q}_2\otimes...\otimes \hat{Q}_{m+1}$), we then write $h_i= w + \varepsilon_i$
and we optimise w.r.t. $\Psi$ to find $\hat{w}_{i+1}$.

\textbf{Bounds of interest.} We present two possible choices of pairs $(\Psi,\Phi)$ derived from the disintegrated results presented in \cref{sec: disintegrated_bounds}. Doing so, we explicit two ready-to-use declinations of \Cref{alg: OPBD_alg}.

\begin{corollary}
  \label{cor: OPBD_optim_funcs}
  For any distribution $\mu$ over $\mathcal{Z}^m$, any online predictive sequences of Gaussian measures with fixed variance $\hat{Q}_i= \mathcal{N}(\hat{w}_i, \sigma^2\mathbf{I}_d)$ and $P_i= \mathcal{N}(w_i^0,\sigma^2\mathbf{I}_d)$, any $\lambda>0$, w.p. $1-\delta$ over $S\sim\mu$ and $(h_i=\hat{w}_{i+1} + \varepsilon_i)_{i=1..m}\sim \hat{Q}_2\otimes...\otimes \hat{Q}_{m+1}$,
  the bound of \cref{eq: OPBD_train_bound} holds for the two following pairs $\Psi,\Phi$:

\begin{align}
  \label{eq: rivasplata_OPBD}
     \Psi_{\normalfont1}(h_i,\hat{w}_{i+1}, w_i^0) & = \frac{1}{\lambda}\left( \frac{||\hat{w}_{i+1} + \varepsilon_i- w_i^0||^2 - ||\varepsilon||^2}{2\sigma^2} \right) \quad \Phi_{\normalfont1}(m) = \frac{\lambda m K^2}{2} + \frac{\log(1/\delta)}{\lambda},\\
    \label{eq: viallard_OPBD}
    \Psi_{\normalfont2}(h_i, \hat{w}_{i+1}, w_i^0)) &  = \frac{1}{\lambda}\frac{||\hat{w}_{i+1}-w_i^0||^2}{2\sigma^2} \quad \Psi_{\normalfont2}(m) = \lambda m K^2 + \frac{3\log(1/\delta)}{2\lambda}.
  \end{align}

  Where the notation $\normalfont1, \normalfont2$ denote whether the functions have been derived from adapted theorems of \citealp{rivasplata2020pac,viallard2021general} recalled in \cref{sec: disintegrated_bounds}
  We then can use \cref{alg: OPBD_alg} with  \cref{eq: rivasplata_OPBD}, \cref{eq: viallard_OPBD}.

\end{corollary}

Proof is deferred to \cref{sec: proofs_sec4}. Note that in \cref{cor: OPBD_optim_funcs}, we identified $\hat{Q}_i$ to $\hat{w}_i$ and for the last formula, $\Psi$ has no dependency on $h_i$.

\textbf{Comparison with \cref{eq: pacb_online_alg}.} The main difference with \cref{eq: pacb_online_alg} provided by the disintegrated framework is that the optimisation route does not include an expected term within the optimisation objective. The main advantage is a weaker computational cost when we restrict to Gaussian distributions. The main weakness is a lack of stability as our algorithm now depends at time $i$ on $\ell(h+\varepsilon_i,z_i)$ so on $\varepsilon_i$ directly.
We denote that \cref{eq: rivasplata_OPBD} is less stable than \cref{eq: viallard_OPBD} as it involves another dependency on $\varepsilon_i$ through $\Psi$.
The reason is that \cite{rivasplata2020pac} proposed a bound involving a disintegrated KL divergence while \cite{viallard2021general} proposed a result involving a Rényi divergence avoiding a dependency on $\varepsilon_i$. We refer to \cref{sec: disintegrated_bounds} for a detailed statement of those properties.

\textbf{Comparison with \cite{hoeven18a}.} Theorem 3 of \citet{hoeven18a} recovers OGD from the exponential weights algorithm by taking a sequence of moving distributions being Gaussians with fixed variance which is exactly what we consider here. From these, they retrieve the classical OGD algorithm as well as its classical convergence rate. Let us compare our results with theirs.

First, if we fix a single step $\eta$ in their bound and assume two traditional assumptions for OGD (a finite diameter $D$ of the convex set and an uniform bound $G$ on the loss gradients), we recover for the OGD (greedy GD in \citealp{hoeven18a}) a rate of $ \frac{D^2}{2\sigma^2\eta} + \frac{\eta\sigma^2TG^2}{2}$. This is, up to constants and notation changes, exactly our $\Psi_i$ ($i\in\{1,2\}$).
Also, we notice a difference in the way to use Gaussian distributions: Theorem 3 of \citet{hoeven18a} is based on their Lemma 1 which provides guarantees for the expected regret. This is a clear incentive to consider as predictors the mean of the sucessive Gaussians of interest. On the contrary, \cref{cor: OPBD_optim_funcs} involves a supplementary level of randomness by considering predictors $h_i$ drawn from our Gaussians. This additional randomness appears in our optimisation process (\cref{alg: OPBD_alg}).
Finally, notice that \citet{hoeven18a} based their whole work on the use of a KL divergence  while \cref{cor: OPBD_optim_funcs} not only exploit a disintegrated KL ($\Psi_1$) but also a Rényi $\alpha$-divergence ($\Psi_2$). Note that we propose a result only for $\alpha=2$ for the sake of space constraints but any other value of $\alpha$ leads to another optimisation objective to explore.

\textbf{OPBD test bounds.}
Similarly to what we did in \cref{sec: online_pacb_procedure}, we also provide \emph{OPBD test bounds} to provide efficiency guarantees for online predicitve sequences (e.g. the output of \cref{alg: OPBD_alg}). Our proposed bounds have the following general form.

For any online predictive sequence $\hat{Q}$, any $\lambda>0$ w.p. $1-\delta$ over $S$ and $(h_1,...,h_{m})\sim \hat{Q}_1\otimes...\otimes \hat{Q}_{m}$:
\begin{align}
  \label{eq: OPBD_test_bound}
  \sum_{i=1}^m  \mathbb{E}[\ell(h_i,z_i) \mid \mathcal{F}_{i-1}]   \leq \sum_{i=1}^m  \ell(h_i,z_i)   + \Phi(m),
\end{align}

with $\Phi$ being a real-valued function(possibly dependent on $\lambda,\delta$ though it is not explicited here).

Note that our predictors $(h_1,...,h_m)$ are now drawn from $\hat{Q}_1\otimes...\otimes \hat{Q}_{m}$. Thus, the left-hand side of the bound considers a $h_i$ drawn from an $\mathcal{F}_{i-1}$-measurable distribution evaluated on $\ell(.,z_i)$: this is effectively a measure of the prediction performance.

We now state a corollary which gives disintegrated guarantees for any online predicitve sequence.

\begin{corollary}
  \label{cor: OPBD_test_bound}
  For any distribution $\mu$ over $\mathcal{Z}^m$, any $\lambda>0$, and any online predictive sequence $(\hat{Q}_i)$, the following holds with probability $1-\delta$ over the sample $S\sim \mu$ and the predictors $(h_1,...,h_m)\sim \hat{Q}_1\otimes...\otimes \hat{Q}_m$, the bound of \cref{eq: OPBD_test_bound} holds with  :

  \begin{align*}
    \Phi_{\normalfont1}(m)= \frac{\lambda m K^2}{2} + \frac{\log(1/\delta)}{\lambda}, \quad \Phi_{\normalfont2}(m) = 2\lambda m K^2 + \frac{\log(1/\delta)}{\lambda}.
  \end{align*}
  Where the notation $\normalfont1, \normalfont2$ denote whether the functions have been derived from adapted theorems of \citealp{rivasplata2020pac,viallard2021general} recalled in \cref{sec: disintegrated_bounds}. The optimised $\lambda$ gives in both cases a $O(\sqrt{m\log(1/\delta)}$.
\end{corollary}

 Proof is deferred to \cref{sec: proofs_sec4}.

\section{Experiments}
\label{sec: experiments}

We adapt the experimental framework introduced in \citet[Sec.5]{cherief2019generalization} to our algorithms (anonymised code available \href{https://anonymous.4open.science/r/Online-PAC-Bayes-learning-044F}{here}). We conduct experiments on several real-life datasets, in classification and
linear regression. Our objective is twofold: check the convergence of our learning methods and compare their efficiencies with classical algorithms. We first introduce our experimental setup.

\textbf{Algorithms.}
We consider four online methods of interest: the OPB algorithm of \cref{eq: pacb_online_alg_specific_case} which update through time a Gibbs posterior. We instantiate it with two different priors $\hat{Q}_1$: a Gaussian distribution and a Laplace one. We also implement \Cref{alg: OPBD_alg} with the functions $\Psi_1,\Psi_2$ from \cref{cor: OPBD_optim_funcs}.
To assess efficiency, we implement the classical OGD (as described in \citealp[Alg. 1 of][]{zinkevich2003online}) and the SVB method of \citet{cherief2019generalization}.

\textbf{Binary Classification.} At each round $i$ the learner receives a data point $x_{i} \in \mathbb{R}^{d}$ and predicts its label $y_{i} \in\{-1,+1\}$ using $\left\langle x_{i}, h_{i}\right\rangle$, with $h_i= \mathbb{E}_{h\sim \hat{Q}_i}[h]$ for OPB methods or $h_i$ being drawn under $\hat{Q}_i$ for OPBD methods.
The adversary reveals the true value $y_{i}$, then the learner suffers the loss $\ell(h_i,z_i)=\left(1-y_{i} h_{i}^{T} x_{i}\right)_{+}$ with $z_i=(x_i,y_i)$ and $a_{+}=a$ if $a>0$ and $a_{+}=0$ otherwise. This loss is unbounded but can be thresholded.

\textbf{Linear Regression.} At each round $i$, the learner receives a set of features $x_{i} \in \mathbb{R}^{d}$ and predicts $y_{i} \in \mathbb{R}$ using $\left\langle x_{i}, h_{i}\right\rangle$ with $h_i= \mathbb{E}_{h\sim \hat{Q}_i}[h]$ for SVB and OPB methods or $h_i$ being drawn under $\hat{Q}_i$ for OPBD methods.
Then the adversary reveals the true value $y_{t}$ and the learner suffers the loss $\ell(h_i,z_i)=\left(y_{i}-h_{i}^{T} x_{i}\right)^{2}$ with $z_i=(x_i,y_i)$. This loss is unbounded but can be thresholded.

\textbf{Datasets.} We consider four real world dataset: two for classification (Breast Cancer and Pima Indians), and two for regression (Boston Housing and California Housing). All datasets except the Pima Indians have been directly extracted from \texttt{sklearn} \citep{scikit-learn}.
Breast Cancer dataset \citep{street1993nuclear} is available \href{https://archive.ics.uci.edu/ml/datasets/Breast+Cancer+Wisconsin+(Diagnostic)}{here} and comes from the UCI ML repository  as well as the Boston Housing dataset \citep{belsley2005regression} which can be obtained \href{https://archive.ics.uci.edu/ml/machine-learning-databases/housing/}{here}. California Housing dataset \citep{pace1997sparse} comes from the StatLib repository and is available \href{https://www.dcc.fc.up.pt/~ltorgo/Regression/cal_housing.html}{here}.
Finally, Pima Indians dataset \citep{smith1988using} has been recovered from this Kaggle \href{https://www.kaggle.com/datasets/uciml/pima-indians-diabetes-database}{repository}.  Note that we randomly permuted the observations to avoid to learn irrelevant human ordering of data (such that date or label).

\textbf{Parameter settings.}
 We ran our experiments on a 2021 MacBookPro with an M1 chip and 16 Gb RAM.
 For OGD, the initialisation point is $\mathbf{0}_{\mathbb{R}^d}$ and the values of the learning rates are set to $\eta=1 / \sqrt{m}$.
 For SVB, mean is initialised to $\mathbf{0}_{\mathbb{R}^d}$ and covariance matrix to $\text{Diag}(1)$. Step at time $i$ is $\eta_i= 0.1/\sqrt{i}$.
 For both of the OPB algorithms with Gibbs posterior, we chose $\lambda= 1/m$. As priors, we took respectively a centered Gaussian vector with the covariance matrix $\text{Diag}(\sigma^2)$ ($\sigma=1.5$) and an iid vector following the standard Laplace distribution.
 For the OPBD algorithm with $\Psi_{\normalfont1}$, we chose $\lambda = 10^{-4}/m$, the initial mean is $\mathbf{0}_{\mathbb{R}^d}$ and
 our fixed covariance matrix is $\text{Diag}(\sigma^2)$ with $\sigma= 3.10^{-3}$.
 For the OPBD algorithm with $\Psi_{\normalfont1}$, we chose $\lambda = 2.10^{-3}/m$, the initial mean is $\mathbf{0}_{\mathbb{R}^d}$ and our covariance matrix is $\text{Diag}(\sigma^2)$ with $\sigma= 10^{-2}$.  The reason of those higher scale parameters and variance is that  $\Psi$ from \citet{rivasplata2020pac} is more stochastic (yet unstable) than the one \citet{viallard2021general}.

\begin{figure}
   \centering
   \begin{subfigure}[b]{0.45\textwidth}
     \centering
     \includegraphics[width=\textwidth]{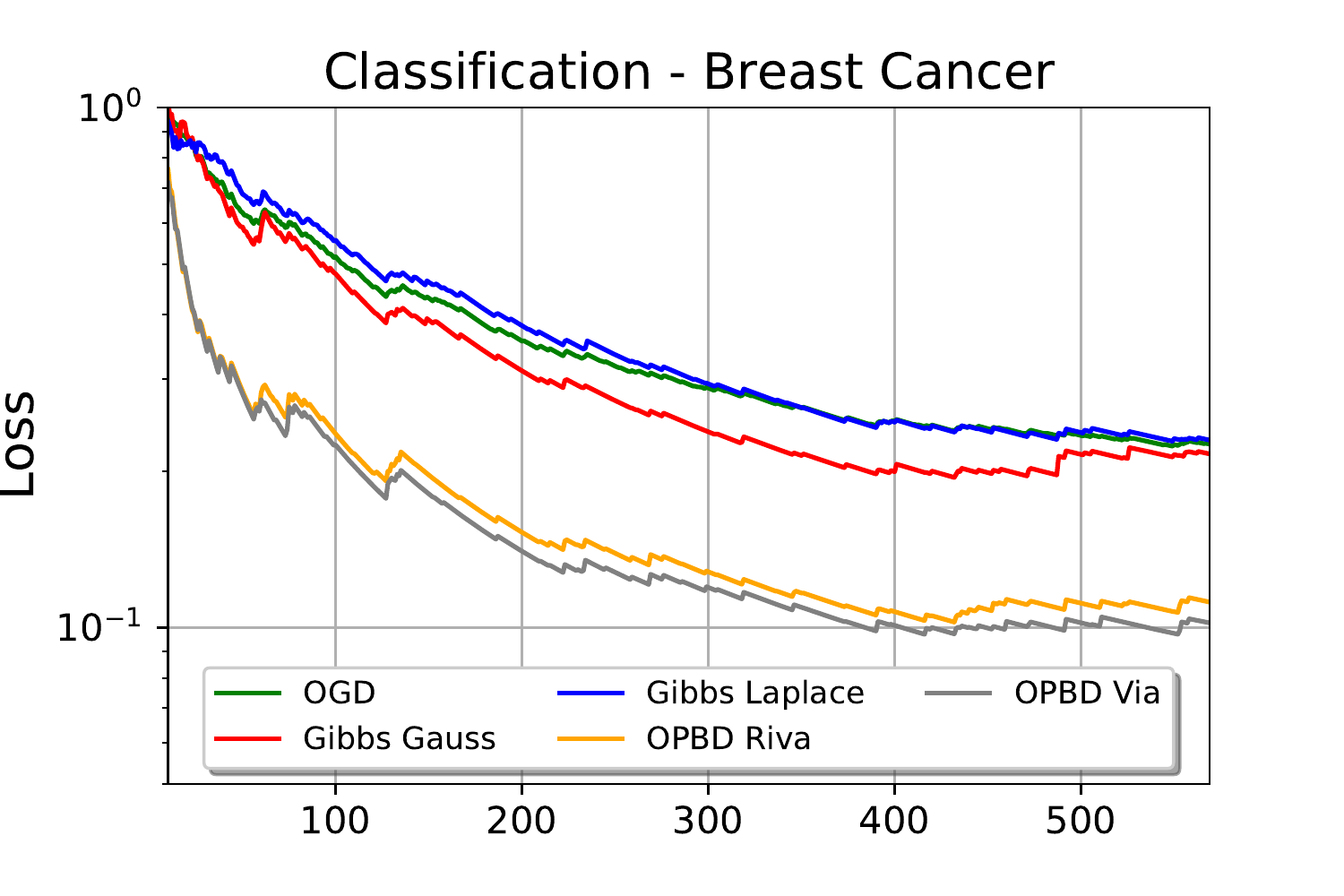}
  \end{subfigure}~
  \begin{subfigure}[b]{0.45\textwidth}
    \centering
    \includegraphics[width=\textwidth]{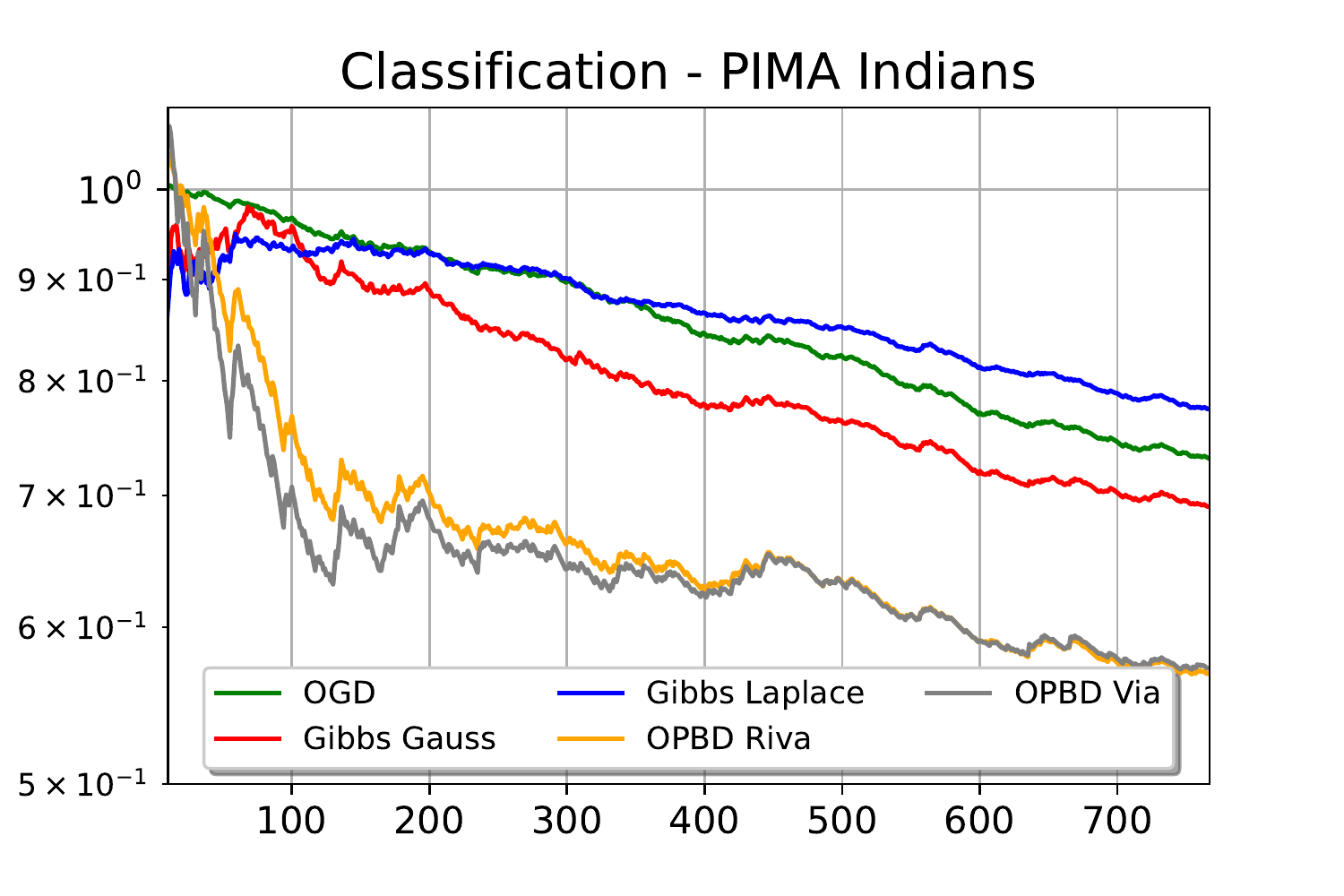}
 \end{subfigure}\\
 \begin{subfigure}[b]{0.45\textwidth}
   \centering
   \includegraphics[width=\textwidth]{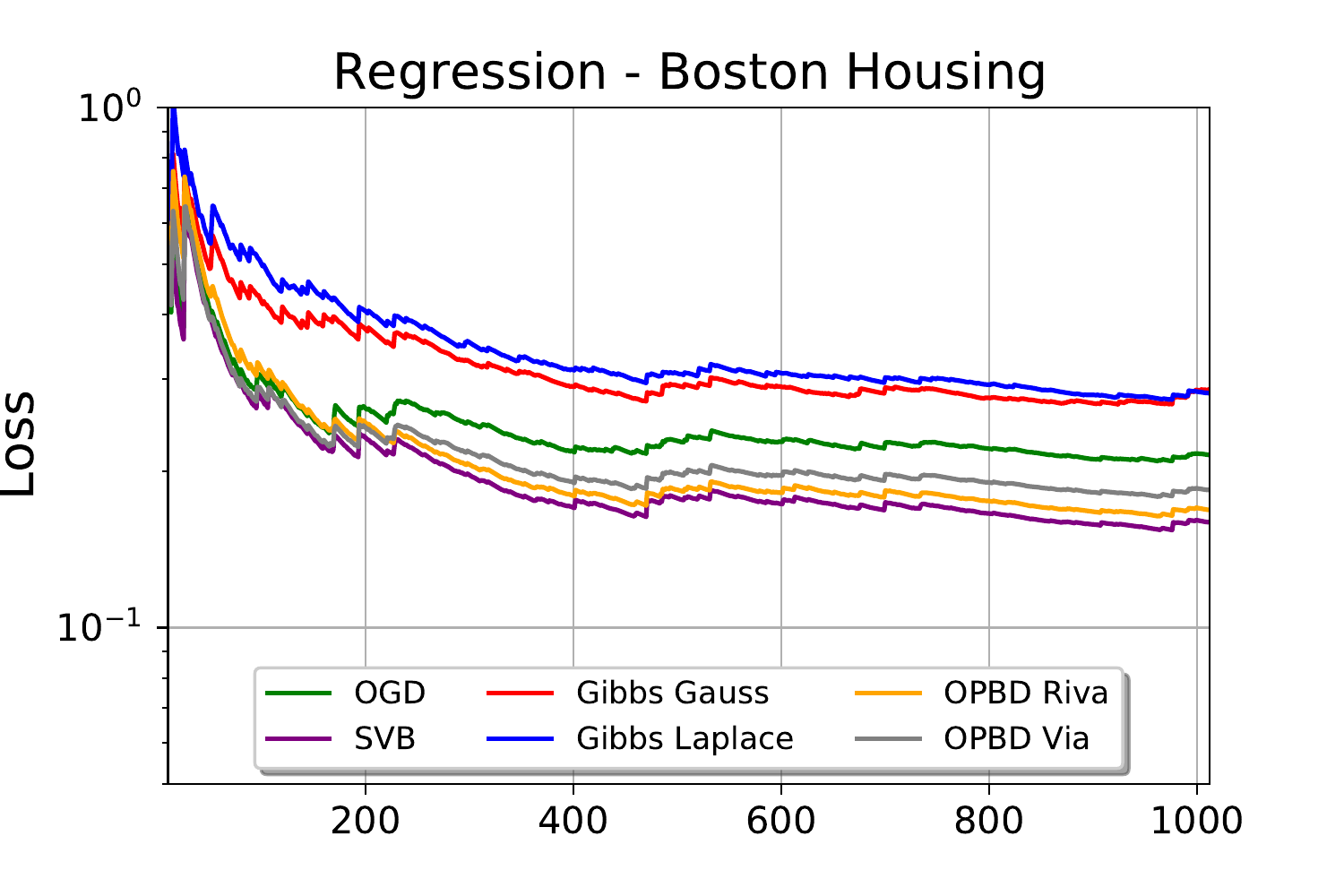}
\end{subfigure}~
\begin{subfigure}[b]{0.45\textwidth}
  \centering
  \includegraphics[width=\textwidth]{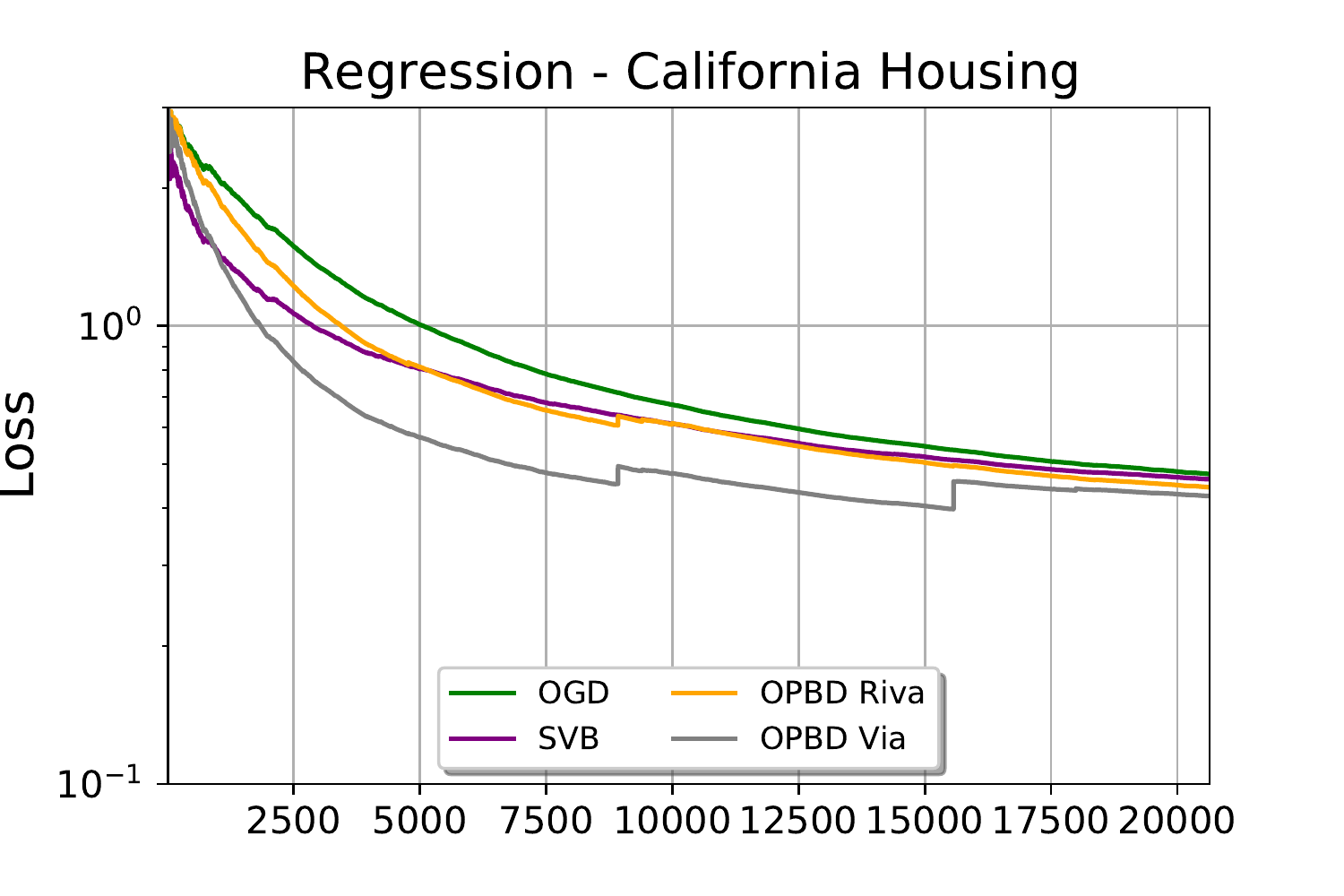}
\end{subfigure}
\caption{Averaged cumulative losses for all four considered datasets. 'Gibbs Gauss' denotes OPB with Gaussian Prior, 'Gibbs Laplace' denotes OPB with Laplace prior. 'OPBD Riva' denotes OPBD with $\Psi_{\normalfont1}$, 'OPBD Via' denotes OPBD with $\Psi_{\normalfont2}$. }
\label{fig: exp_results}
 \end{figure}

 \textbf{Experimental results.}
 For each dataset, we plot the evolution of the average cumulative loss $\sum_{i=1}^{t} \ell\left(h_{i},z_i\right) / t$ as a function of the step $t=1, \ldots, m$, where $m$ is the dataset size and $h_{i}$ is the decision made by the learner $h_i$ at step $i$. The results are gathered in \cref{fig: exp_results}

\textbf{Empirical findings.} OPB with Gaussian prior ('Gibbs Gauss') outperforms OGD on all datasets except California Housing (on which this method is not implemented ) while OPB with Laplace prior ('Gibbs Laplace') always fail w.r.t. OGD. OPB methods fail to compete with SVB on the Boston Housing dataset. OPBD methods compete with SVB on regression problems and clearly outperforms OGD on classification tasks. OPBD with $\Psi_{\normalfont2}$ (labeled as 'OPBD Via' in \cref{fig: exp_results}) performs better on the California Housing dataset while OPBD with $\Psi_{\normalfont1}$ (labeled as 'OPBD Riva') is more efficient on the Boston Housing dataset. Both methods performs roughly equivalently on classification tasks.  This brief experimental validation shows the consistency of all our online procedures as we observe a visible decrease of the cumulative losses through time. It particularly shows that OPBD procedures improve on OGD on these dataset. We refer to \cref{sec: error_bars} for additional table gathering the error bars of our OPBD methods.

\textbf{Why do we perform better than OGD?} As stated in \cref{sec: OPBD_procedure}, OGD can be recovered as a Gaussian approximation of the exponential weights algorithm (EWA). Thus, a legitimate question is why do we perform better than OGD as our OPBD methods are also based on a Gaussian surrogate of EWA?  \cite{hoeven18a} only used Gaussians distributions with fixed variance as a technical tool when the considered predictors are the Gaussian means. In our work, we exploited a richer characteristic of our distributions in the sense our predictors are points sampled from our Gaussians and not only the means. This also has consequences in our learning algorithm as at time $i$ of our \cref{alg: OPBD_alg}, our optimisation step involves a noise $\varepsilon_i\sim \mathcal{N}(0,\sigma^2\mathbf{I})$. Thus, we believe that OPBD methods should perform at least as well as OGD.
We write 'at least' as we think that the higher flexibility due to this additional level of randomness might result in slightly better empirical performances, as seen on the few datasets in \cref{fig: exp_results}.

\section{Conclusion}

We establish links between Online Learning and PAC-Bayes. We show that PAC-bayesian bounds are useful to derive new OL algorithms. We also prove sound theoretical guarantees for such algorithms. We emphasise that all of our results stand for any general bounded loss, especially no convexity assumption is needed.
Having no convexity assumption on the loss paves the way to exciting future practical studies, starting with \emph{Spiking Neural Network} which is investigated in an online fashion (see \citealp{lobo2020spiking} for a recent survey). A follow-up question on the theoretical part is whether we can relax the bounded loss assumption: we leave this for future work.

\section*{Acknowledgements}

We thank several anonymous reviewers as well as the Area Chair of the Neurips comitee who provided insigthful comments and suggestions which greatly enhanced the quality of our comparisons with literature and of the discussions around our theorems.

\bibliography{biblio}

\clearpage
\section*{Checklist}

\begin{enumerate}

\item For all authors...
\begin{enumerate}
  \item Do the main claims made in the abstract and introduction accurately reflect the paper's contributions and scope?
    \answerYes{}
  \item Did you describe the limitations of your work?
    \answerYes{ From the abstract to the conclusion, we explicited our main assumption (bounded loss).}
  \item Did you discuss any potential negative societal impacts of your work?
    \answerNA{Due to the theoretical nature of our contribution, we do not foresee immediate societal impact. If anything, we certainly hope a better theoretical understanding of online learning algorithms can foster a more responsible use by practitionners.}
  \item Have you read the ethics review guidelines and ensured that your paper conforms to them?
    \answerYes{}
\end{enumerate}

\item If you are including theoretical results...
\begin{enumerate}
  \item Did you state the full set of assumptions of all theoretical results?
    \answerYes{for all theorems and corollaries}
        \item Did you include complete proofs of all theoretical results?
    \answerYes{But deferred to \cref{sec: proofs} due to space constraints.}
\end{enumerate}

\item If you ran experiments...
\begin{enumerate}
  \item Did you include the code, data, and instructions needed to reproduce the main experimental results (either in the supplemental material or as a URL)?
    \answerYes{We have included the url in \cref{sec: experiments}, note that this is an anonymous repository.}
  \item Did you specify all the training details (e.g., data splits, hyperparameters, how they were chosen)?
    \answerYes{ All details are gathered in \cref{sec: experiments} so that all readers can replicate our experiments and validate our fundings.}
        \item Did you report error bars (e.g., with respect to the random seed after running experiments multiple times)?
    \answerNo{We chose to consider the classical online gradient descent as main comparison.}
        \item Did you include the total amount of compute and the type of resources used (e.g., type of GPUs, internal cluster, or cloud provider)?
    \answerYes{We precised the machine we used in the 'Parameters Settings' paragraph of \cref{sec: experiments}.}
\end{enumerate}

\item If you are using existing assets (e.g., code, data, models) or curating/releasing new assets...
\begin{enumerate}
  \item If your work uses existing assets, did you cite the creators?
    \answerYes{ We precised the origin of our datasets in the 'Datasets' paragraph of \cref{sec: experiments}.}
  \item Did you mention the license of the assets?
    \answerNA{}
  \item Did you include any new assets either in the supplemental material or as a URL?
    \answerYes{All the original code is available on the anonymised repository precised in \cref{sec: experiments}.}
  \item Did you discuss whether and how consent was obtained from people whose data you're using/curating?
    \answerNA{}
  \item Did you discuss whether the data you are using/curating contains personally identifiable information or offensive content?
    \answerNA{}
\end{enumerate}

\item If you used crowdsourcing or conducted research with human subjects...
\begin{enumerate}
  \item Did you include the full text of instructions given to participants and screenshots, if applicable?
    \answerNA{}
  \item Did you describe any potential participant risks, with links to Institutional Review Board (IRB) approvals, if applicable?
    \answerNA{}
  \item Did you include the estimated hourly wage paid to participants and the total amount spent on participant compensation?
    \answerNA{}
\end{enumerate}

\end{enumerate}

\newpage
\appendix

\section{Background}

\subsection{Reminder on Online Gradient Descent }
\label{sec: OGD_reminder}

For the sake of completeness we re-introduce the projected Online Gradient Descent (OGD) on a convex set $\mathcal{K}$. This is a first example of online learning philosophy. It may be the algorithm that applies to the most general setting
of online convex optimization. This algorithm,
which is based on standard gradient descent from offline optimization, was
introduced in its online form by \cite{zinkevich2003online}.
In each iteration, the algorithm takes a step from the previous point in
the direction of the gradient of the previous cost. This step may result in
a point outside of the underlying convex set. In such cases, the algorithm
projects the point back to the convex set, i.e. finds its closest point in the
convex set. We precise this algorithm works with the assumptions of a convex set $\mathcal{K}$ bounded in diameter by $D$ and of bounded gradients (by a certain $G$).
We also assume here to have a dataset $S=(z_t)_{t=1..T}$ and to be coherent with the online learning philosophy, we assume that for each $t>0$, we possess a loss function $\ell_t$ depending on the points $(z_1,...z_t)$. We present OGD in \cref{alg: gradient_descent}

\begin{algorithm}[ht]
 \SetAlgoLined
 \SetKwInOut{Initialisation}{Initialisation}
 \SetKwInOut{Parameter}{Parameters}
 \Parameter{Epoch T, step-size $(\eta)$ }
 \Initialisation{Convex set $\mathcal{K}$, Initial point $\theta_0 \in\mathcal{K}$, T, step sizes $(\eta_t)_t$ }
\For{ each iteration $t$ in $1..T$}{ Compute $f'(\theta_{n}) $\\
Play (observe) $\theta_t$ and compute the cost $f_t(\theta_t)$
Update and project \[ \zeta_{t} = \theta_{t-1} - \eta \nabla \ell_t(\theta_{t-1})  \]
\[ \theta_t = \Pi_\mathcal{K}(\zeta_t)  \]}
\textbf{Return} $\theta_{T}$
 \caption{Projected OGD onto a convex $\mathcal{K}$ with fixed step $\eta$.}
 \label{alg: gradient_descent}
 \end{algorithm}

One now defines the notion of regret which is the classical quantity to evaluate the performance of an online algorithm.
 \begin{definition}
One defines the \emph{regret} of a decision sequence $(\theta_t)$ at time $T$  w.r.t. a point $\theta$ as:

\[ Regret_T(\theta):= \sum_{t=1}^T \ell_t(\theta_t) -  \sum_{t=1}^T \ell_t(\theta)  \]

\end{definition}

Now we state a regret bound which can be found in \cite[Eq 2.5]{shalev2012online} although we slightly modified the result, which uses additional hypotheses from \cite{hazan2016introduction}.

\begin{proposition}
  \label{prop: OGD_bound}
  Assume that $\mathcal{K}$ has a fixed diameter $D$ and that the gradients of any point is bounded by $G$. Then for any $\theta\in\mathcal{K}$, the regret of projected OGD with fixed step $\eta$ satisfies:

  \[ Regret_T(\theta) \leq \frac{D^2}{2\eta} + \eta T G^2     \]
\end{proposition}

\subsection{About PAC-Bayes learning}

We provide here more information about PAC-Bayes learning. We propose a local framework; then expose what APC-Bayes lerning ais to do and finally state two celebrated PAC-Bayes theorems.

\paragraph{An usual framework}
We first state the following framework:
\begin{itemize}
  \item $\mathcal{H}$ is a space of considered predictors
  \item $\mathcal{Z}$ is a data space. $z$ can be an unlabeled data $x$ or a couple $(x,y)$ of a point with its label. We assume that $\mu$ is a distribution over $\mathcal{Z}$ which rules the distribution of our data.
  \item $\ell: \mathcal{H}\times \mathcal{Z}\rightarrow \mathbb{R}^+$ is a loss function i.e. the learning objective we want to minimise.
  \item $S=(z_1,...z_m)$ an iid dataset following $\mu$.
  \item The generalisation risk for $h\in\mathbb{H}$: $R(h)= \mathbb{E}_{z\sim \mu}[\ell(h,z)]$.
  \item The empirical risk $R_m(h)= \frac{1}{m}\sum_{i=1}^m \ell(h,z_i)$.
\end{itemize}

\paragraph{What is PAC-Bayes learning?}
PAC-Bayes learning is about learning a meaningful data-dependent posterior distribution $Q$ from a (classically data-free) prior $P$ without necessarily exploiting the Bayes formula. PAC-Bayes learning controls the \emph{expected generalisation error}:
\[  \mathbb{E}_{h\sim Q}[R(h)] := \mathbb{E}_{h\sim Q}\mathbb{E}_{z\sim \mu} [\ell(h,z)], \]
 which is the averaged error that would make a predictor drawn from our posterior distribution on a new point (usually this objective holds under an iid assumption on our dataset).

\paragraph{Two classical theorems.} We state below two celebrated PAC-Bayesian results: the McAllester bound enriched with Maurer's remark as stated in \cite[Thm.1]{guedj2019primer} and Catoni's bound (\citealp[Thm 1.2.6]{catoni2007pac}). Those theorems both holds with the assumptions of iid data and loss bounded by $1$. Note that in \cref{th: naive_pac_bayes}, we also proposed another bound which is a corollary of Catoni's one.

\begin{theorem}[McAllester' bound]
  For any prior distribution $P$, we have with probability $1-\delta$ over the $m$-sample $S$, for any posterior distribution $Q$ such that $Q\ll P$:

\[  \mathbb{E}_{h\sim Q}\left[ R(h)\right] \leq \mathbb{E}_{h\sim Q}\left[ R_m(h)\right] + \sqrt{\frac{KL(Q,P) + \log(2\sqrt{m}/\delta)}{2m}},    \]
where $KL$ is the Kullback-Leibler divergence.

\end{theorem}

\begin{theorem}
  For any prior distribution $P$, any $\lambda>0$, we have with probability $1-\delta$ over the $m$-sample $S$, for any posterior distribution $Q$ such that $Q\ll P$:

\[  \mathbb{E}_{h\sim Q}\left[ R(h)\right] \leq \frac{1-\exp \left\{-\frac{\lambda \mathbb{E}_{h\sim Q}\left[ R_m(h)\right]}{m}-\frac{KL(Q || P)-\log (\delta)}{m}\right\}}{1-\exp \left(-\frac{\lambda}{m}\right)},    \]
where $KL$ is the Kullback-Leibler divergence.

\end{theorem}

\section{Discussion about \cref{th: main_thm_online}}
\label{sec: discussion_main_thm}

\subsection{Comparison with classical PAC-Bayes}

\label{sec: comparison_main_thm}

The goal of this section is to show how good \cref{th: main_thm_online} compared to a naive approach which consists in applying classical PAC-Bayes results sequentially. The interest of this section is twofold:

\begin{itemize}
  \item First, presenting a classical PAC-Bayes result extracted and adapted from \cite{alquier2016variational} which is formally close to what we propose.
  \item Second, showing that a naive (yet natural) approach to obtain online PAC-Bayes bound leads to a deteriorated bound.
\end{itemize}

We first state our PAC-Bayes bound of interest.

\begin{theorem}[Adapted from  \cite{alquier2016variational}, Thm 4.1]
  \label{th: naive_pac_bayes}
  Let $S=(z_1,...,z_m)$ be an iid sample from the same law $\mu_0$.
  For any data-free prior $P$, for any loss function $\ell$ bounded by $K$, any $\lambda>0,\delta\in ]0;1[$, one has with probability $1-\delta$ for any posterior $Q\in\mathcal{M}_1(\mathcal{H})$:

  \[ \mathbb{E}_{h\sim Q}\mathbb{E}_{z\sim \mu_0}[\ell(h,z)] \leq \frac{1}{m} \sum_{i=1}^m \mathbb{E}_{h\sim Q}[\ell(h,z_i)] + \frac{\operatorname{KL}(Q\| P) + \log(1/\delta)}{\lambda} + \frac{\lambda K^2}{2m} \]
\end{theorem}

\begin{remark} Two remarks about this result:

  \begin{itemize}
    \item \cref{th: naive_pac_bayes} is a particular case of the original theorem from \cite{alquier2016variational} as we take the case of a bounded loss which implies the subgaussianity of the random variables $\ell(.,z_i)$ and then allows us to recover the factor $\frac{\lambda K^2}{m}$
    \item This theorem is derived from \cite{catoni2007pac} and constitutes a good basis to compare ourselves with as it similar formally similar.
  \end{itemize}
\end{remark}

\paragraph{Naive approach} A naive way to obtain OPB bounds is to apply $m$ times \cref{th: naive_pac_bayes} (one per data) on batches of size $1$ and then summing up the associated bounds. Thus one has the benefits of classical PAC-Bayes bound without having no more the need of data-free priors nor the iid assumption. The associated result is stated below:

\begin{theorem}
  \label{th: naive_approach}
  For any distributions $\mu_1,...,\mu_m$ over $\mathcal{Z}$ (such that $z_i\sim \mu_i$), any $\lambda>0$ and any online predictive sequence (used as priors) $(P_i)$, the following holds with probability $1-\delta$ over the sample $S\sim\mu$ for any posterior sequence $(Q_i)$ :

  \begin{align*}
    \sum_{i=1}^m \mathbb{E}_{h_i\sim Q_{i}}\left[ \mathbb{E}_{z_i\sim \mu_i}[\ell(h_i,z_i)]    \right] \leq \sum_{i=1}^m \mathbb{E}_{h_i\sim Q_{i}}\left[ \ell(h_i,z_i) \right] +
    \frac{\operatorname{KL}(Q_{i}\| P_i)}{\lambda} + \frac{\lambda m K^2}{2} + \frac{m\log(m/\delta)}{\lambda}.
  \end{align*}
\end{theorem}

Recall that here again we assimilate the stochastic kernels $Q_i, P_i$ to the data-dependent distributions $Q_i(S,.), P_i(S,.)$

\begin{proof}
  First of all, for any $i$, we apply \cref{th: naive_pac_bayes} $m$ to the batch $\{ z_i\}$. This allows us to consider $P_i$ as a prior as it does not depend on the current data. We then have, taking $\delta'=\delta/m$, for any $i\in\{1..m\}$ with probability $ 1- \delta/m$:

  \[ \mathbb{E}_{h_i\sim Q_{i}}\left[ \mathbb{E}_{z_i\sim \mu_i}[\ell(h_i,z_i)]    \right] \leq  \mathbb{E}_{h_i\sim Q_{i}}\left[ \ell(h_i,z_i) \right] +
  \frac{\operatorname{KL}(Q_{i}\| P_i)}{\lambda} + \frac{\lambda K^2}{2} + \frac{\log(m/\delta)}{\lambda} \]

  Then, taking an union bound on those $m$ events ensure us that with probability $1-\delta$, for any $i\in\{1..m\}$:

  \[ \mathbb{E}_{h_i\sim Q_{i}}\left[ \mathbb{E}_{z_i\sim \mu_i}[\ell(h_i,z_i)]    \right] \leq  \mathbb{E}_{h_i\sim Q_{i}}\left[ \ell(h_i,z_i) \right] +
  \frac{\operatorname{KL}(Q_{i}\| P_i)}{\lambda} + \frac{\lambda  K^2}{2} + \frac{\log(m/\delta)}{\lambda} \]

  Finally, summing those $m$ inequalities ensure us the final result with probability $1-\delta$.

\end{proof}

\paragraph{Comparison between \cref{th: main_thm_online} and \cref{th: naive_approach}}   Three points are noticeable between those two theorems:

\begin{itemize}
  \item First of all, the main issue with \cref{th: naive_approach} is that has a strongly deteriorated rate of $O\left(\frac{m\log(m/\delta)}{\lambda} \right)$ instead of the rate in  $O\left(\frac{\log(1/\delta)}{\lambda} \right)$ proposed in \cref{th: main_thm_online}.
  More precisely, the problem is that we do not have a sublinear bound: one cannot ensure any learning through time.
  This point justifies the need of the heavy machinery exploited in \cref{th: main_thm_online} proof as it allows a tighter convergence rate.
  \item The second point point lies in the controlled quantity on the left hand-side of the bound. \cref{th: naive_approach} controls $A:=\sum_{i=1}^m \mathbb{E}_{h_i\sim Q_{i}}\left[ \mathbb{E}_{z_i\sim \mu_i}[\ell(h_i,z_i)]    \right]$
  instead of $B:=\sum_{i=1}^m \mathbb{E}_{h_i\sim Q_{i}}\left[ \mathbb{E}[\ell(h_i,z_i) \mid \mathcal{F}_{i-1}]    \right]$.

  $A$ is a less dynamic quantity than $B$ in the sense that it does not imply any evolution through time, it just considers global expectations. Doing so, $A$ does not take into account that at each time step we have acces to all te past data to predict the future, this may explain the deteriorated convergence rate. Thus $B$, which appears to be a suitable quantity to control to perform online PAC-Bayes (see \cref{sec: deeper_analysis_main_thm} for additional explanations)

  \item Finally, an interesting point is that in \cref{th: naive_approach} the bound, while looser, holds unformly for any posterior sequence contrary to \cref{th: main_thm_online} which holds only for a specific posterior sequence. This point will have a consequence for optimisation. We will come back later on this in \cref{sec: main_thm_and_optim}.
\end{itemize}

\subsection{A deeper analysis of \cref{th: main_thm_online}}
\label{sec: deeper_analysis_main_thm}

This section includes discussion about our proof technique and why all the assumptions made are necessary. We also propose a short discussion about the benefits and limitations of an online PAC-Bayesian framework as well as a deeper reflexion about the new term our bound introduce.

\paragraph{Why do we need an online predictive sequence as priors? }

This condition is fully exploited when dealing with the exponential moment $\xi_m$ in the proof (see \cref{l: exp_moment_online} proof). Indeed, the fact of having $P_i$ being $\mathcal{F}_{i-1}$-measurable is essential to apply conditional Fubini (\cref{l: cond_fubini}). Note that the condition $\forall i , P_{i-1}\ll P_{i}$ is not necessary as the weaker condition $\forall i, P_1 \ll P_i$ would suffice here.
However, note that when we particularise our theorem, for instance if we choose in \cref{cor: online_procedure} $P_i= \hat{Q_i}$, one recovers the condition $\hat{Q}_{i}\ll\hat{Q}_{i+1}$ to have finite KL divergences. Hence the interest of taking directly an online predictive sequence.

\paragraph{About the boundedness assumption}

The only moment where we invoke the boundedness assumption is in \cref{l: exp_moment_online}'s proof where we apply the conditionnal Hoeffding lemma. This lemma actually translates that the sequence of r.v. $(\ell(.,z_i)_{i=1..m}$ is \emph{conditionally subgaussian} wrt the past i.e for any $i$, $h_i\in\mathcal{H}; \lambda\in\mathbb{R}$:

\[ \mathbb{E}[\exp(\lambda \Tilde{\ell}_i(h_i,z_i)) \mid \mathcal{F}_{i-1}] \leq \exp\left( \frac{\lambda^2K^2}{2}\right)\]

 where $\Tilde{\ell}_i(h_i,z_i)= \mathbb{E}[\ell(h_i,z_i)\mid \mathcal{F}_{i-1}]-  \ell(h_i,z_i)$.

 This condition is the one truly involved in our heavy machinery. However, we chose to restrict ourselves to the stronger assumption of bounded loss function for the sake of clarity. However, an interesting open direction is to find whether there exists concrete classes of unbounded losses which may satisfy either conditional subgaussianity or others conditions (such as conditional Bernstein condition for instance).

 \paragraph{Reflections about the left hand side of \cref{th: main_thm_online}.}

 We study in this paragraph the following term
  $$B:=\sum_{i=1}^m \mathbb{E}_{h_i\sim Q_{i}}\left[ \mathbb{E}[\ell(h_i,z_i) \mid \mathcal{F}_{i-1}]    \right]$$ has naturally arisen in our work as the right term to compare our empirical loss with to perform the conditional Hoeffding lemma.
 Taking a broader look, we now interpret this term as the right quantity to control if one wants to perform online PAC-Bayes learning. Indeed this term is a 'best of both world' quantity bridging PAC-Bayes and online learning:

 \begin{itemize}
   \item From the PAC-Bayes point of view one keeps the control on average (cf the conditional expectation in $B$) on a novel data drawn at each time step. This point is crucial in the PAC-Bayes literature as our posteriors are designed to generalise well to unseen data.
   \item From the Online Learning point of view, one keeps the control of a sequence of points generated from an online algorithm. Because an online learning algorithm generate a prediction for future points while having access to past data, the conditional expectation in $B$ translates this.
 \end{itemize}

Finally this conditional expectation appears to be a good tradeoff between the classical expectation on data appearing in the PAC-Bayes literature (see e.g. \cref{th: naive_pac_bayes}) and the local control that we have in online learning by only dealing with the performance of a sequence of points generated from a learning algorithm (see e.g. \cref{prop: OGD_bound})

\paragraph{About the interest of an Online PAC-Bayesian framework}
The main shift our work does with classical online learning literature is that it does not consider the celebrated regret but instead focuses on $B$ which is a cumulative expected loss conditionned to the past. This shift does not invalidate our work but put some relief to hte guarantees Online PAC-Bayes learning can provide that Online Learning cannot and reversely.

\begin{itemize}
  \item Online PAC-Bayes ensure a good potential for generalisation as it deals with the control of conditional expectation. This can be useful if one wants to deal with a periodic process for instance.
  \item Online Learning through the regret compares the studied sequence of predictors (typically generated from an online learning algorithm) and tries to compare it to the best fixed strategy (static regret) or the best dynamic one (dynamic regret). In this way, OL algorithms want to ensure that their predictions are closed from the optimal solution. This point is not guaranteed by our online PAC-Bayesian study.
  \item However the limitations of online learning can arise if the studied problem has a huge variance (for instance micro-transactions in finance). In this case those algorithms can follow an unpredictable optimisation route while PAC-Bayes still ensure a good performance on average (knowing the past) in this case.
  \item Finally, we want to emphasize that PAC-Bayesian learning circumvent a problem of \emph{memoryless learning} which appears in classical OL algorithms. For instance, the OGD algorthm (see \cref{sec: OGD_reminder}) uses once a data and do not memorise it for further use. This problem does not happen in Online PAC-Bayes learning. Indeed, we take the example of the procedure \cref{eq: pacb_online_alg_specific_case} which generates Gibbs posterior which keep in mind the influence of past data.
\end{itemize}

\subsection{\cref{th: main_thm_online} and optimisation}
\label{sec: main_thm_and_optim}

In this section we discuss about the way Thm 2.2 can be thought in the framework of an optimisation process as we did in \cref{sec: online_pacb_procedure,sec: OPBD_procedure}.

\paragraph{A significant change compared to classical PAC-Bayes}

\cref{th: main_thm_online} holds 'for any posterior sequence $(Q_i)$ the following holds with probability $1-\delta$ over the sample $S\sim\mu$ ' while most classical PAC-Bayesian results such that \cref{th: naive_pac_bayes} holds 'with probability $1-\delta$ over the sample $S\sim\mu$ for any posterior $Q$'. This change is significant as our theorem does not control simultaneaously all possible sequences of posteriors but only holds for one.
Thus, \cref{th: main_thm_online} has to be seen as a local or pointwise theorem and not as a global one. In classical PAC-Bayes, this local behavior is a brake on the optimisation process. But as we develop below, it is not the case in our online framework.

\paragraph{\cref{th: main_thm_online} is compatible with online optimisation}

We first recall that classically, an online algorithm like OGD (see \cref{sec: OGD_reminder}) performs one optimisation step per arriving data. Thus, at time $m$, such algorithm will perform $m$ optimisation steps and generate $m$ predictors. Similarly the OPB algorithm of \cref{eq: pacb_online_alg} generates $m$ distribution in $m$ time steps.

We insist on the fact that, \cref{th: main_thm_online} \textbf{and all its corollaries throughout our paper are valid for a sequence of $m$ posteriors and not only a single one.} A key point is that whatever the number $m$ of data, our theoretical guarantee wil still be valid for $m$ posterior distributions with the approximation term $\log(1/\delta)$ (and not $\log(m/\delta)$ as an union bound would provide for a classical PAC-Bayes theorem).

For this reason, given an online PAC-Bayes algorithm, \cref{th: main_thm_online} is suited for optimisation. Indeed, having a bound valid for a sequence of posteriors ensures guarantees for a single run of our OPB algorithm. This point is crucial to bridge a link with online learning as regret bounds (e.g. \cref{prop: OGD_bound}) also provide guarantees for a single sequence of predictors. In online learning however, those guarantees are mainly deterministic (because based on convex optimisation properties) but not totally: the recent work of \cite{wintenberger2021stochastic} proposed PAC regret bounds for its general Stochastic Online Convex Optimisation framework.

An interesting open challenge is to overcome the pointwise behavior of our theorem, for that, we need to rethought \cite[Thm 2.1]{rivasplata2020pac} as this basis is pointwise itself. Given we consider a sequence of data-dependent priors one cannot apply the classical change of measure inequality to ensure guarantees holding uniformly on posterior sequences.

\paragraph{A crucial point: having an explicit OPB/OPBD algorithm}

In our previous paragraph we said that our bound were suitable for optimisation given an OPB/OPBD algorithm. We now provide some precision about this point. All the procedures provided in the paper (i.e. \cref{eq: pacb_online_alg}, \cref{alg: OPBD_alg}) take into account an update phase implying an argmin. Luckily for our procedures, this argmin is explicit:
\begin{itemize}
  \item For the OPB algorithm of \cref{eq: pacb_online_alg}, the argmin is solved thanks to the variational formulation of the Gibbs posterior
  \item For OPBD algorithms, given the explicit choices of $\Psi$ given in \cref{cor: OPBD_optim_funcs}, argmin becomes explicit when one has a derivable loss function.
\end{itemize}

In both cases, this explicit argmin ensure our procedure of interest generates explictly a single posterior per time step: we have a well-defined sequence of $m$ posteriors at time $m$.
Doing so the guarantees of \cref{th: main_thm_online} holds for this sequence.

\section{A reminder on PAC-Bayesian disintegrated bounds}
\label{sec: disintegrated_bounds}

We present two PAC-Bayesian disintegrated bounds valid with data-dependent priors (i.e. any stochastic kernels).
\begin{itemize}
  \item The first one is Th. 1) i) from \cite{rivasplata2020pac} which provides a disintegrated version of \cref{th: main_thm_online}.
  \item The second one is Thm 2. from \cite{viallard2021general} which involves Rényi divergence instead of the classical $KL$. Note that this bound has originally been stated for data-indepedent prior, which is why we revisit the proof to adapt it to the stochastic kernel framework.
\end{itemize}

\begin{proposition}[Th 1) i) \cite{rivasplata2020pac}]
\label{prop: rivasplata_disintegrated}
 Let $P \in \mathcal{M}_{1}(\mathcal{S})$, $Q^{0} \in \texttt{Stoch}(\mathcal{Z}^m,\mathcal{F})$. Let $f: \mathcal{S} \times \mathcal{H} \rightarrow \mathbb{R}$ be any measurable function.
Then for any $Q \in \texttt{Stoch}(\mathcal{Z}^m,\mathcal{F})$ and any $\delta \in(0,1)$, with probability at least $1-\delta$ over the random draw of $S \sim P$ and $h\sim Q_S$, we have:
$$
f(S,h) \leq \log\left(\frac{dQ_{S}}{dQ_{S}^{0}}(h) \right)+\log (\xi_m / \delta) .
$$

where $\xi_m:=\int_{\mathcal{S}} \int_{\mathcal{H}} e^{f(s, h)} Q_{s}^{0}(d h) P(d s)$ and $\frac{dQ_{S}}{dQ_{S}^{0}}$ is the Radon Nykodym derivative of $Q_S$ w.r.t. $Q_S^0$.
\end{proposition}

\begin{proposition}[Adapted from Th. 2 of \cite{viallard2021general}]
  \label{prop: viallard_disintegrated}
   Let $\mu \in \mathcal{M}_{1}(\mathcal{S})$, $Q^{0} \in \texttt{Stoch}(\mathcal{Z}^m,\mathcal{F})$. Let  $\alpha>1$ and  $f: \mathcal{S} \times \mathcal{H} \rightarrow \mathbb{R}^+$ be any measurable function.

   Then for any $Q \in \texttt{Stoch}(\mathcal{Z}^m,\mathcal{F})$ such that for any $S\in\mathcal{Z}^m, Q_S>>Q_S^0,\; Q_S^0>>Q_S $ and any $\delta \in(0,1)$, with probability at least $1-\delta$ over the random draw of $S \sim \mu$ and $h\sim Q_S$, we have:

   \[\frac{\alpha}{\alpha-1} \log (f(S,h)) \leq \frac{2 \alpha-1}{\alpha-1} \log \frac{2}{\delta}+D_{\alpha}\left(Q_{S} \| Q_S^0\right)+
   \log \left(\underset{S^{\prime} \sim \mu}{\mathbb{E}} \underset{h^{\prime} \sim Q_{S'}^0}{\mathbb{E}}
   f\left(S^{\prime},h^{\prime}\right)^{\frac{\alpha}{\alpha-1}}\right) \]

   where $D_\alpha(Q\|P)= \frac{1}{\alpha-1}\log\left( \mathbb{E}\left[ \mathbb{E}_{h\sim P} \left(\frac{dQ}{dP}(h)\right)^\alpha   \right] \right)$ is the Rényi diverence of order $\alpha$.

\end{proposition}

Note that Viallard et al. original bound only stand for data-free priors and i.i.d data. However it appears their proof works with any stochastic kernel as prior and any distribution over the dataset. We propose below an adaptation of their proof below to fit with those more general assumptions.

\subsection{Proof of \cref{prop: viallard_disintegrated} }

\begin{proof}[Proof]
 For any sample $S$ and any stochastic kernel $Q$, note that $f(S,h)$ is a non-negative random variable. Hence, from Markov's inequality we have
$$
\underset{h \sim Q_{S}}{\mathbb{P}}\left[f(S,h) \leq \frac{2}{\delta} \underset{h^{\prime} \sim Q_{S}}{\mathbb{E}} f\left( S,h^{\prime}\right)\right] \geq 1-\frac{\delta}{2} \Longleftrightarrow
\underset{h \sim Q_{S}}{\mathbb{E}} \mathds{1}\left[f(S,h) \leq \frac{2}{\delta} \underset{h^{\prime} \sim Q_{S}}{\mathbb{E}} f\left(S,h'\right)\right] \geq 1-\frac{\delta}{2}
$$
Taking the expectation over $S \sim \mu$ to both sides of the inequality gives

\begin{multline*}
\underset{S \sim \mu}{\mathbb{E}}\; \underset{h \sim Q_{S}}{\mathbb{E}} \mathds{1}\left[f(S,h) \leq \frac{2}{\delta} \underset{h^{\prime} \sim Q_{S}}{\mathbb{E}} f(S,h')\right] \geq 1-\frac{\delta}{2} \\
\Longleftrightarrow
\underset{S \sim \mu, h \sim Q_{S}}{\mathbb{P}}\left[f(S,h) \leq \frac{2}{\delta} \underset{h^{\prime} \sim Q_{S}}{\mathbb{E}} f(S,h')\right] \geq 1-\frac{\delta}{2}.
\end{multline*}

Taking the logarithm to both sides of the equality and multiplying by $\frac{\alpha}{\alpha-1}>0$, we obtain
$$
\underset{S \sim \mu, h \sim Q_{S}}{\mathbb{P}}\left[\frac{\alpha}{\alpha-1} \log (f(S,h)) \leq \frac{\alpha}{\alpha-1} \log \left(\frac{2}{\delta} \underset{h^{\prime} \sim Q_{S}}{\mathbb{E}} f(S,h')\right)\right] \geq 1-\frac{\delta}{2} .
$$
We develop the right side of the inequality in the indicator function and make the expectation of the hypothesis over $Q_S^0$ our "prior" stochadtic kernel appears. Indeed, because for any $S\in\mathcal{S}, Q_S>> Q_S^0$ and $Q_S^0>> Q_S$ one can write properly $\frac{dQ_S}{dQ_S^0}$ and $ \frac{dQ_S^0}{dQ_S} = \left( \frac{dQ_S}{dQ_S^0}\right)^{-1}$ the Radon-Nykodym derivatives. Thus we have

\begin{multline*}
 \frac{\alpha}{\alpha-1} \log \left(\frac{2}{\delta} \underset{h^{\prime} \sim Q_{S}}{\mathbb{E}} f(S,h')\right) \\ =\frac{\alpha}{\alpha-1} \log \left(\frac{2}{\delta} \underset{h^{\prime} \sim Q_{S}}{\mathbb{E}} \frac{dQ_S}{dQ_S^0}(h')\frac{dQ_S^0}{dQ_S}(h') f(S,h') \right) \\
 =\frac{\alpha}{\alpha-1} \log \left(\frac{2}{\delta} \underset{h^{\prime} \sim Q_S^0}{\mathbb{E}} \frac{dQ_S}{dQ_S^0}(h') f(S,h')\right) .
\end{multline*}

Remark that $\frac{1}{r}+\frac{1}{s}=1$ with $r=\alpha$ and $s=\frac{\alpha}{\alpha-1}$. Hence, we can apply Hölder's inequality:
$$
\underset{h^{\prime} \sim Q_S^0}{\mathbb{E}} \frac{dQ_S}{dQ_S^0}(h') f(S,h') \leq\left[\underset{h^{\prime} \sim Q_S^0}{\mathbb{E}}\left(\frac{dQ_S}{dQ_S^0}(h')\right)^{\alpha}\right]^{\frac{1}{\alpha}}\left[\underset{h^{\prime} \sim Q_S^0}{\mathbb{E}} f(S,h')^{\frac{\alpha}{\alpha-1}}\right]^{\frac{\alpha-1}{\alpha}} .
$$
Then, by taking the logarithm; adding $\log \left(\frac{2}{\delta}\right)$ and multiplying by $\frac{\alpha}{\alpha-1}>0$ to both sides of the inequality, we obtain

\begin{multline*}
\frac{\alpha}{\alpha-1} \log \left(\frac{2}{\delta} \underset{h^{\prime} \sim Q_S^0}{\mathbb{E}} \frac{dQ_S}{dQ_S^0}(h') f(S,h')\right) \\
\leq \frac{\alpha}{\alpha-1} \log \left(\frac{2}{\delta}\left[\underset{h^{\prime} \sim Q_S^0}{\mathbb{E}}\left(\frac{dQ_S}{dQ_S^0}(h')\right)^{\alpha}\right]^{\frac{1}{\alpha}}\left[\underset{h^{\prime} \sim Q_S^0}{\mathbb{E}} f(S,h')^{\frac{\alpha}{\alpha-1}}\right]^{\frac{\alpha-1}{\alpha}}\right) \\
 =\frac{1}{\alpha-1} \log \left(\underset{h^{\prime} \sim Q_S^0}{\mathbb{E}}\left[\frac{dQ_S}{dQ_S^0}(h')\right]^{\alpha}\right)+\frac{\alpha}{\alpha-1} \log \frac{2}{\delta}+\log \left(\underset{h^{\prime} \sim Q_S^0}{\mathbb{E}} f(S,h')^{\frac{\alpha}{\alpha-1}}\right) \\
=D_{\alpha}\left(Q_{S} \| Q_S^0\right)+\frac{\alpha}{\alpha-1} \log \frac{2}{\delta}+\log \left(\underset{h^{\prime} \sim Q_S^0}{\mathbb{E}} f(S,h')^{\frac{\alpha}{\alpha-1}}\right)
\end{multline*}

From this inequality, we can deduce that
\begin{multline}
  \label{eq: viallard_dis_eq11}
  \underset{S \sim \mu, h \sim Q_{S}}{\mathbb{P}}\left[\frac{\alpha}{\alpha-1} \log (f(S,h)) \leq D_{\alpha}\left(Q_{S} \| Q_S^0\right)+\frac{\alpha}{\alpha-1} \log \frac{2}{\delta}+\log \left(\underset{h^{\prime} \sim Q_S^0}{\mathbb{E}} f(S,h')^{\frac{\alpha}{\alpha-1}}\right)\right] \\
  \geq 1-\frac{\delta}{2} .
\end{multline}

Note that $\mathbb{E}_{h^{\prime} \sim Q_S^0} f(S,h')^{\frac{\alpha}{\alpha-1}}$ is a non-negative random variable, hence, we apply Markov's inequality to have
$$
\underset{S \sim \mu}{\mathbb{P}}\left[\underset{h^{\prime} \sim Q_S^0}{\mathbb{E}} f(S,h')^{\frac{\alpha}{\alpha-1}} \leq \frac{2}{\delta} \underset{S^{\prime} \sim \mu}{\mathbb{E}} \underset{h^{\prime} \sim Q_S^0}{\mathbb{E}} f(S',h')^{\frac{\alpha}{\alpha-1}}\right] \geq 1-\frac{\delta}{2} .
$$
Since the inequality does not depend on the random variable $h \sim Q_{S}$, we have

\begin{multline*}
\underset{S \sim \mu}{\mathbb{P}}\left[\underset{h^{\prime} \sim Q_S^0}{\mathbb{E}} f(S,h')^{\frac{\alpha}{\alpha-1}} \leq \frac{2}{\delta} \underset{S^{\prime} \sim \mu}{\mathbb{E}} \underset{h^{\prime} \sim Q_S^0}{\mathbb{E}} f(S',h')^{\frac{\alpha}{\alpha-1}}\right] \\
=\underset{S \sim \mu}{\mathbb{E}}\mathds{1}\left[\underset{h'\sim Q_S^0}{\mathbb{E}} f(S,h')^{\frac{\alpha}{\alpha-1}} \leq \frac{2}{\delta} \underset{S^{\prime} \sim \mu}{\mathbb{E}} \underset{h^{\prime} \sim Q_S^0}{\mathbb{E}} f(S',h')^{\frac{\alpha}{\alpha-1}}\right] \\
=\underset{S \sim \mu}{\mathbb{E}}\; \underset{h \sim Q_{S}}{\mathbb{E}} \mathds{1}\left[\underset{h^{\prime} \sim Q_S^0}{\mathbb{E}} f(S,h')^{\frac{\alpha}{\alpha-1}} \leq \frac{2}{\delta} \underset{S^{\prime} \sim \mu}{\mathbb{E}} \underset{h^{\prime} \sim Q_S^0}{\mathbb{E}} f(S',h')^{\frac{\alpha}{\alpha-1}}\right] \\
=\underset{S \sim \mu, h \sim Q_{S}}{\mathbb{P}}\left[\underset{h^{\prime} \sim Q_S^0}{\mathbb{E}} f(S,h')^{\frac{\alpha}{\alpha-1}} \leq \frac{2}{\delta} \underset{S^{\prime} \sim \mu}{\mathbb{E}} \underset{h^{\prime} \sim Q_S^0}{\mathbb{E}} f(S',h')^{\frac{\alpha}{\alpha-1}}\right] .
\end{multline*}

Taking the logarithm to both sides of the inequality and adding $\frac{\alpha}{\alpha-1} \log \frac{2}{\delta}$ give us

\begin{multline}
\underset{S \sim \mu, h \sim Q_{S}}{\mathbb{P}}\left[\underset{h^{\prime} \sim Q_S^0}{\mathbb{E}} f(S,h')^{\frac{\alpha}{\alpha-1}} \leq \frac{2}{\delta} \underset{S^{\prime} \sim \mu}{\mathbb{E}} \underset{h^{\prime} \sim Q_S^0}{\mathbb{E}} f(S',h')^{\frac{\alpha}{\alpha-1}}\right] \geq 1-\frac{\delta}{2} \quad \Longleftrightarrow \\
\label{eq: viallard_dis_eq12}
\underset{S \sim \mu, h \sim Q_{S}}{\mathbb{P}}\left[\frac{\alpha}{\alpha-1} \log \frac{2}{\delta}+\log \left(\underset{h^{\prime} \sim Q_S^0}{\mathbb{E}} f(S,h')^{\frac{\alpha}{\alpha-1}}\right) \leq \right. \\
 \left. \frac{2 \alpha-1}{\alpha-1} \log \frac{2}{\delta}+\log \left(\underset{S^{\prime} \sim \mu}{\mathbb{E}} \underset{h^{\prime} \sim Q_S^0}{\mathbb{E}} f(S',h')^{\frac{\alpha}{\alpha-1}}\right)\right] \geq 1-\frac{\delta}{2} .
\end{multline}

Combining Equation \cref{eq: viallard_dis_eq11} and \cref{eq: viallard_dis_eq12} with a union bound gives us the desired result.
\end{proof}

\section{Proofs}
\label{sec: proofs}

\subsection{Proof of \cref{th: main_thm_online}}

\label{sec: proof_main_thm_online}

\paragraph{Background} We first recall \cite[Thm 2]{rivasplata2020pac}.

\begin{theorem}
\label{th: rivasplata2020}
Let $\mu \in \mathcal{M}_{1}(\mathcal{S})$, $Q^{0} \in \texttt{Stoch}(\mathcal{S},\mathcal{F})$. Let $k$ be a positive integer,  any  $A: \mathcal{S} \times \mathcal{H} \rightarrow \mathbb{R}^{k}$ a measurable function and $F: \mathbb{R}^{k} \rightarrow \mathbb{R}$ be a convex function .
Then for any $Q \in \texttt{Stoch}(\mathcal{S},\mathcal{F})$ and any $\delta \in(0,1)$, with probability at least $1-\delta$ over the random draw of $S \sim \mu$ we have
$$
F\left(Q_{S}\left[A_{S}\right]\right) \leq \mathrm{KL}\left(Q_{S} \| Q_{S}^{0}\right)+\log (\xi_m / \delta) .
$$

where $\xi_m:=\int_{\mathcal{S}} \int_{\mathcal{H}} e^{f(s, h)} Q_{s}^{0}(d h) P(d s)$ and $Q_S[A_S]:= Q_S[A(S,.)]= \int_{\mathcal{H}} A(S,h) Q_S(dh)$.
\end{theorem}

\begin{proof}[Proof of \cref{th: main_thm_online}]

To fully exploit the generality of \cref{th: rivasplata2020}, we aim to design a $m$-tuple of probabilities. Thus, our predictor set of interest is $\mathcal{H}_m:= \mathcal{H}^{\otimes m}$ and then, our predictor $h$ is a tuple $(h_1,..,h_m)\in\mathcal{H}$. Throughout our study, our stochastic kernels $Q,Q^0$ will belong to the specific class $\mathcal{C}$ defined below:

\begin{align}
  \label{eq: class_kernels}
  \mathcal{C}:= \left\{ Q \mid   \exists (Q_i)_{i=1..m}\text{s.t.}\; \forall S,\;  Q(S,.) = Q_1(S,.)\otimes...\otimes Q_m(S,.)      \right\}
\end{align}

\noindent Thus our kernels are such that conditionally to a given sample, our predictors $(h_1,...,h_m)$ are drawn independently.

\noindent We now apply \cref{th: rivasplata2020}. To do so, we consider the following function $A: \mathcal{S}\times \mathcal{H}_m \rightarrow \mathbb{R}^2$ such that $\forall S= (z_i)_{i=1..m},h= (h_i)_{i=1..m}\in \mathcal{S}\times \mathcal{H}_m$:

\[  A(S,h)= \left(\sum_{i=1}^m \mathbb{E}[\ell(h_i,z_i)\mid \mathcal{F}_{i-1}], \sum_{i=1}^m \ell(h_i,z_i)  \right)   \]

\noindent $A$ is indeed measurable in both of its variables. For a fixed $\lambda>0$, we set the function $F$ to be $F(x,y)= \lambda(x-y)$ .

 \noindent The only thing left to set up is our stochastic kernels. To do so, let $P=(P_1,...P_m)$ be an online predictive sequence, we then define $Q^0\in\mathcal{C}$ (defined in \cref{eq: class_kernels}) s.t. for any sample $S$, $Q^0_S = P_1(S,.)\otimes...\otimes P_m(S,.)$. We also fix $Q_1,...,Q_m$ to be any (posterior)
 stochastic kernels and similarly we define the stochastic kernel $Q\in\mathcal{C}$ such that for any sample $S$, $Q(S,.) = Q_1(S,.)\otimes...\otimes Q_m(S,.)$.

 From now, we fix a dataset $S$ and, for the sake of clarity, we assimilate in what follows the stochastic kernels $Q_i,P_i$ to the data-dependent distributions $Q_i(S,.), P_i(S,.)$ (i.e. we drop the dependency in $S$).

\noindent Under those choices, one has:

\begin{multline*}
  Q_S[A_S]  = \int_{h\in\mathcal{H}_m} A(S,h) Q_S(dh_1,...,dh_m) \\
     = \left( \int_{h\in\mathcal{H}_m} \sum_{i=1}^m \mathbb{E}[\ell(h_i,z_i)\mid \mathcal{F}_{i-1}]Q_S(dh_1,...,dh_m), \int_{h\in\mathcal{H}_m}\sum_{i=1}^m \ell(h_i,z_i)  Q_S(dh_1,...,dh_m)    \right)
\end{multline*}
Furthermore, $Q\in\mathcal{C}$, thus $Q_S(dh_1,...,dh_m)= \Pi_{i=1}^m Q_i(dh_i)$ so:`
\begin{align*}
  Q_S[A_S] &= \left(  \sum_{i=1}^m \mathbb{E}_{h_i\sim Q_i}[\mathbb{E}\left[\ell(h_i,z_i)\mid \mathcal{F}_{i-1}]\right], \sum_{i=1}^m \mathbb{E}_{h_i\sim Q_i}[\ell(h_i,z_i)]       \right)
\end{align*}

Finally:
\begin{align*}
      F(Q_S[A_S])&= \lambda \left(\sum_{i=1}^m \mathbb{E}_{h_i\sim Q_i}[\mathbb{E}\left[\ell(h_i,z_i)\mid \mathcal{F}_{i-1}]\right]- \sum_{i=1}^m \mathbb{E}_{h_i\sim Q_i}[\ell(h_i,z_i)] \right)
\end{align*}

Applying \cref{th: rivasplata2020} and re-organising the terms gives us with probability $1-\delta$:

\begin{align*}
\sum_{i=1}^m \mathbb{E}_{h_i\sim Q_i}\left[\mathbb{E}[\ell(h_i,z_i)\mid \mathcal{F}_{i-1}]\right] & \leq \sum_{i=1}^m \mathbb{E}_{h_i\sim Q_i}[\ell(h_i,z_i)] + \frac{KL(Q_S\|Q^0_S)}{\lambda} + \frac{\log(\xi_m/\delta)}{\lambda}
\end{align*}
\noindent Thus:
\begin{align}
  \label{eq: temp_result_online_thm}
 \sum_{i=1}^m \mathbb{E}_{h_i\sim Q_i}\left[\mathbb{E}[\ell(h_i,z_i)\mid \mathcal{F}_{i-1}]\right] \leq \sum_{i=1}^m \mathbb{E}_{h_i\sim Q_i}[\ell(h_i,z_i)] + \sum_{i=1}^m \frac{KL(Q_i\|P_i)}{\lambda} + \frac{\log(\xi_m/\delta)}{\lambda}
\end{align}

\noindent The last line holding because for a fixed $S$, $Q_S= Q_1\otimes...\otimes Q_m$ and $Q_S^0= P_1\otimes...\otimes P_m$.

\noindent The last term to control is
\begin{align*}
\xi_m = \mathbb{E}_S\left[\mathbb{E}_{h_1,...,h_m \sim Q^0_S} \left[\exp\left(\lambda \sum_{i=1}^m \Tilde{\ell}_i(h_i,z_i)\right)\right]\right]
\end{align*}
with $\Tilde{\ell}_i(h_i,z_i)= \mathbb{E}[\ell(h_i,z_i)\mid \mathcal{F}_{i-1}]-  \ell(h_i,z_i)$. Hence the following lemma.

\begin{lemma}
\label{l: exp_moment_online}
One has for any $m$, $\xi_m \leq \exp\left(\frac{\lambda^2m K^2}{2}\right)$ with $K$ bounding $\ell$.
\end{lemma}

The proof of this lemma is deferred to \cref{sec: proof_exp_moment_online}

To conclude the proof, we just bound $\xi_m$ by the result of \cref{l: exp_moment_online} within \cref{eq: temp_result_online_thm}.
\end{proof}

\subsubsection{Proof of \cref{l: exp_moment_online}}
\label{sec: proof_exp_moment_online}

\begin{proof}[Proof of \cref{l: exp_moment_online}]
  We prove our result by recursion: for $m=1$, $S={z_1}$ and one knows that $P_1$ is $\mathcal{F}_0$ measurable yet it does not depend on $S$. Thus for any $h_1\in\mathcal{H}$, $\mathbb{E}[\ell(h_1,z_1)\mid \mathcal{F}_{0}]= \mathbb{E}[\ell(h_1,z_1)]$. We then has:
  \begin{align*}
    \xi_1 &= \mathbb{E}_{S}\mathbb{E}_{h_1\sim P_1} [\Tilde{\ell}_1(h_1,z_1)]\\
    &= \mathbb{E}_{h_1\sim P_1} \mathbb{E}_{S} [\Tilde{\ell}_1(h_1,z_1)] & \text{by Fubini} \\
    & \leq \exp{\frac{\lambda^2K^2}{2}}
  \end{align*}

The last line  holding because for any $h_1\in\mathcal{H}$, $\Tilde{\ell}_1(h_1,z_1)$ is a centered variable belonging in $[-K,K]$ a.s. and so one can apply Hoeffding's lemma to conclude.

\noindent Assume the result is true at rank $m-1\geq 0$. We then has to prove the result at rank $m$. Our strategy consists in conditioning by $\mathcal{F}_{m-1}$ within the expectation over $S$:

\begin{align*}
  \xi_m & = \mathbb{E}_S\left[\mathbb{E}_{h_1,...,h_m \sim Q^0_S} \left[\exp\left(\lambda \sum_{i=1}^m \Tilde{\ell}_i(h_i,z_i)\right)\right]\right]\\
  \intertext{First, we use that $Q^0\in\mathcal{C}$, thus $Q_S^0 = P_1\otimes...\otimes P_m$ (i.e. our data are drawn independently for a given $S$):}
  &= \mathbb{E}_S\left[\Pi_{i=1}^m \mathbb{E}_{h_i \sim P_i} \left[\exp\left(\lambda  \Tilde{\ell}_i(h_i,z_i)\right)\right]\right]\\
  \intertext{We now condition by $\mathcal{F}_{m-1}$ and use that $\Pi_{i=1}^{m-1}\mathbb{E}_{h_i \sim P_i} \left[\exp\left(\lambda  \Tilde{\ell}_i(h_i,z_i)\right)\right]$ is a $\mathcal{F}_{m-1}$-measurable r.v.}
  \xi_m & = \mathbb{E}_S\left[ \Pi_{i=1}^{m-1}\mathbb{E}_{h_i \sim P_i} \left[\exp\left(\lambda  \Tilde{\ell}_i(h_i,z_i)\right)\right] \mathbb{E}
\left[ \mathbb{E}_{h_m\sim P_m} [\exp(\lambda \Tilde{\ell}_m(h_m,z_m))] \mid \mathcal{F}_{m-1} \right] \right]
\end{align*}

Now our next step is to use a variant of Fubini valid for $\mathcal{F}_{m-1}$- measurable measures.

\begin{lemma}[Conditional Fubini]
  \label{l: cond_fubini}
  Let $f: \mathcal{H}\times \mathcal{Z} \rightarrow \mathbb{R}^+$.
  For a $sigma$-algebra $\mathcal{F}$ over $\mathcal{Z}$ and a measure $P$ over $\mathcal{H}$ such that
  \begin{itemize}
    \item $P$ is a $\mathcal{F}$-measurable r.v.
    \item There exists a constant measure (a.s.) $P_0$ such that $P>>P_0$.
  \end{itemize}
  Then one has almost surely, for any r.v. $z$ over $\mathcal{Z}$:

  \[ \mathbb{E}\left[\mathbb{E}_{h\sim P} [f(h,z)] \mid \mathcal{F} \right] =  \mathbb{E}_{h\sim P} \left[ \mathbb{E}[f(h,z)\mid \mathcal{F}] \right]    \]
\end{lemma}

\noindent The proof of this lemma lies at the end of this section.

\noindent We then fix $\mathcal{F}= \mathcal{F}_{m-1}$ and $f(h,z)= \exp(\lambda\Tilde{\ell}_i(h,z))$. Furthermore, because we assumed the sequence $(P_i)$ to be an online predictive sequence, $P_m$ is $\mathcal{F}_{m-1}$-measurable and $P_m>>P_1$ with $P_1$ a data-free prior. One then applies \cref{l: cond_fubini}:

\[ \mathbb{E}
\left[ \mathbb{E}_{h_m\sim P_m} [\exp(\lambda \Tilde{\ell}_m(h_m,z_m))] \mid \mathcal{F}_{m-1} \right] =  \mathbb{E}_{h_m\sim P_m} \left[ \mathbb{E}
[\exp(\lambda \Tilde{\ell}_m(h_m,z_m)) \mid \mathcal{F}_{m-1}] \right] . \]

\noindent Yet, injecting this result onto $\xi_m$ provides:

\begin{align*}
  \xi_m = \mathbb{E}_S\left[ \Pi_{i=1}^{m-1}\mathbb{E}_{h_i \sim P_i} \left[\exp\left(\lambda  \Tilde{\ell}_i(h_i,z_i)\right)\right] \mathbb{E}_{h_m\sim P_m} \left[ \mathbb{E}
  [\exp(\lambda \Tilde{\ell}_i(h_m,z_m)) \mid \mathcal{F}_{m-1}] \right] \right]
\end{align*}

The final remark is to notice that for any $h_m\in\mathcal{H}$, $\mathbb{E}[\Tilde{\ell}_m(h_m,z_m)\mid \mathcal{F}_{m-1}] = 0$ and $\Tilde{\ell}_m(h_m,z_m)\in [-K,K]$ a.s. then one can apply the conditional Hoeffding's lemma which ensure us that for any $\lambda>0$:

\[ \mathbb{E}
[\exp(\lambda \Tilde{\ell}_m(h_m,z_m)) \mid \mathcal{F}_{m-1}] \leq \exp\left( \frac{\lambda^2K^2}{2}   \right). \]

One then has $\xi_m \leq \exp\left( \frac{\lambda^2K^2}{2}   \right) \xi_{m-1}$. The recursion assumption concludes the proof.

\end{proof}

\begin{proof}[Proof of \cref{l: cond_fubini}]
 Let $A$ be a $\mathcal{F}$-measurable event. One wants to show that
  \[ \mathbb{E}\left[\mathbb{E}_{h\sim P} [f(h,z)]\mathds{1}_A \right] = \mathbb{E}\left[\mathbb{E}_{h\sim P} \left[ \mathbb{E}[f(h,z)\mid \mathcal{F}] \right] \mathds{1}_A \right] \]
  \noindent Where the first expectation in each term is taken over $z$. This will be enough to conclude that
  \[ \mathbb{E}\left[\mathbb{E}_{h\sim P} [f(h,z)]\mid \mathcal{F} \right] = \mathbb{E}_{h\sim P} \left[ \mathbb{E}[f(h,z)\mid \mathcal{F}] \right]   \]
  \noindent thanks to the definition of conditional expectation. We first start by using the fact that $P$ is $\mathcal{F}$-measurable and that $P>>P_0$ with $P_0$ a constant measure. This is enough to obtain that the Radon-Nykodym derivative $\frac{dP}{dP_0}$ is a $\mathcal{F}$-measurable function, thus:

  \begin{align*}
     \mathbb{E}\left[\mathbb{E}_{h\sim P} [f(h,z)]\mathds{1}_A \right]&=  \mathbb{E}\left[\mathbb{E}_{h\sim P_0} \left[f(h,z)\frac{dP}{dP_0}(h)\right]\mathds{1}_A(z) \right] \\
     &= \mathbb{E}\left[\mathbb{E}_{h\sim P_0} \left[f(h,z)\frac{dP}{dP_0}(h)\mathds{1}_A(z)\right] \right]
     \intertext{Because $f(h,z)\frac{dP}{dP_0}(h)\mathds{1}_A(z) $ is a positive function, and that $P_0$ is fixed, one can apply the classical Fubini-Tonelli theorem:}
     &= \mathbb{E}_{h\sim P_0} \left[ \mathbb{E}\left[f(h,z)\frac{dP}{dP_0}(h)\mathds{1}_A(z)\right] \right] \\
     \intertext{One now conditions by $\mathcal{F}$ and use the fact that $\frac{dP}{dP_0}, \mathds{1}_A$ are $\mathcal{F}$-measurable:  }
     &= \mathbb{E}_{h\sim P_0} \left[ \mathbb{E}\left[\mathbb{E}\left[f(h,z)\mid \mathcal{F}\right]\frac{dP}{dP_0}(h)\mathds{1}_A(z)\right] \right] \\
     \intertext{We finally re-apply Fubini-Tonelli to re-intervert the expectations: }
     &=  \mathbb{E}\left[ \mathbb{E}_{h\sim P_0}\left[\mathbb{E}\left[f(h,z)\mid \mathcal{F}\right]\frac{dP}{dP_0}(h)\mathds{1}_A(z)\right] \right] \\
     &= \mathbb{E}\left[ \mathbb{E}_{h\sim P}\left[\mathbb{E}\left[f(h,z)\mid \mathcal{F}\right]\mathds{1}_A(z)\right] \right]
  \end{align*}

  \noindent This finally proves the announced results, yet concludes the proof.

\end{proof}

\subsection{Proofs of \cref{sec: OPBD_procedure}}
\label{sec: proofs_sec4}
We prove here \cref{cor: OPBD_optim_funcs} and \cref{cor: OPBD_test_bound}.

\subsubsection{Proof of \cref{cor: OPBD_optim_funcs}}
We fix $\hat{Q},P$ to be online predictive sequences (with $\hat{Q}_1,P_1$ being data-free priors). Recall that we assimilated the stochastic kernels $\hat{Q}_i,P_i$ to the their associated data-dependent sitribution given a sample $S$ $\hat{Q}_i(S,.), P_i(S,.)$.

As in \cref{th: main_thm_online}, our predictor set of interest is $\mathcal{H}_m:= \mathcal{H}^{\otimes m}$ and then, our predictor $h$ is a tuple $(h_1,..,h_m)\in\mathcal{H}$. We consider the stochastic kernel $Q$ belonging to the class $\mathcal{C}$ defined in \cref{eq: class_kernels} such that for any $S\in\mathcal{S}, Q(S,.) = \hat{Q}_2\otimes ... \otimes \hat{Q}_{m+1}$.
Similarly one defines $Q^0\in\mathcal{C}$ such that for any $S\in\mathcal{S}, Q^0(S,.) = P_1\otimes ... \otimes P_{m}$

\paragraph{Proof for $(\Psi_{\normalfont1},\Phi_{\normalfont1})$:}

For $\lambda>0$, we set our function $f$ to be for any dataset $S$ and predictor tuple $h=(h_1,...,h_m)$,
\[f(S,h) = \lambda \left(\sum_{i=1}^m \mathbb{E}\left[\ell(h_i,z_i)\mid \mathcal{F}_{i-1}\right]- \sum_{i=1}^m \ell(h_i,z_i) \right)\]

We then apply \cref{prop: rivasplata_disintegrated} with the function $f$, $Q,Q^0$ defined above. One then has by dividing by $\lambda$ with probability $1-\delta$ over $S\sim \mu$ and $h=(h_1,...,h_m)\sim \hat{Q}_2\otimes ... \otimes \hat{Q}_{m+1}$:

\[ \sum_{i=1}^m  \mathbb{E}[\ell(h_i,z_i) \mid \mathcal{F}_{i-1}]   \leq \sum_{i=1}^m  \ell(h_i,z_i)  + \frac{1}{\lambda}\log\left(\frac{dQ_S}{dQ_S^0}(h_i)\right) + \frac{1}{\lambda} \log(\xi_m) + \frac{\log(1/\delta)}{\lambda} \]

And then using the fact that $S\in\mathcal{S}, Q_S = \hat{Q}_2\otimes ... \otimes \hat{Q}_{m+1}, Q^0_S = P_1\otimes ... \otimes P_{m}$ gives us:

\[ \sum_{i=1}^m  \mathbb{E}[\ell(h_i,z_i) \mid \mathcal{F}_{i-1}]   \leq \sum_{i=1}^m  \ell(h_i,z_i)  + \frac{1}{\lambda}\sum_{i=1}^m \log\left(\frac{d\hat{Q}_{i+1}}{dP_i}(h_i)\right) + \frac{1}{\lambda} \log(\xi_m) + \frac{\log(1/\delta)}{\lambda} \]

with $  \xi_m = \mathbb{E}_S\left[\mathbb{E}_{h_1,...,h_m \sim Q_S} \left[\exp\left(\lambda \sum_{i=1}^m \Tilde{\ell}_i(h_i,z_i)\right)\right]\right]$ and for any $i$,
$ \Tilde{\ell}_i(h_i,z_i) = \mathbb{E}\left[\ell(h_i,z_i)\mid \mathcal{F}_{i-1}\right]-  \ell(h_i,z_i) $

Notice that, because $P$ is an online predictive sequence, then one can apply directly \cref{l: exp_moment_online} to conclude that $\xi_m \leq \exp \left( \frac{\lambda^2K^2m}{2} \right)$.

We also use \cite[Lemma 11]{viallard2021general} which derives the calculation of the disintegrated KL divergence between two Gaussians. One then has for any $i$, with $h_i= \hat{w}_{i+1} + \varepsilon_i$:

\[ \log\left(\frac{d\hat{Q}_{i+1}}{dP_i}(h_i)\right) = \frac{||\hat{w}_{i+1} + \varepsilon_i- w_i^0||^2 - ||\varepsilon||^2}{2\sigma^2} \]

Combining those facts altogether allows us to conclude.

\paragraph{Proof for  $(\Psi_{\normalfont2},\Phi_{\normalfont2})$:}

For $\lambda>0$, we set our function $f$ to be for any dataset $S$ and predictor tuple $(h=h_1,...,h_m)$,
\[f(S,h) = \exp\left(\lambda \left(\sum_{i=1}^m \mathbb{E}\left[\ell(h_i,z_i)\mid \mathcal{F}_{i-1}\right]- \sum_{i=1}^m \ell(h_i,z_i) \right) \right)\]
We take $\alpha=2$ and apply this time \cref{prop: viallard_disintegrated}.
One then has by dividing by $2\lambda$ with probability $1-\delta$ over $S\sim \mu$ and $h=(h_1,...,h_m)\sim \hat{Q}_2\otimes ... \otimes \hat{Q}_{m+1}$:

\[\sum_{i=1}^m  \mathbb{E}[\ell(h_i,z_i) \mid \mathcal{F}_{i-1}]   \leq \sum_{i=1}^m  \ell(h_i,z_i)+   \frac{3}{2\lambda}\log \frac{2}{\delta}+\frac{D_{2}\left(Q_{S} \| Q_S^0\right)}{2\lambda}+ \frac{1}{2\lambda}
\log \left(\underbrace{\underset{S^{\prime} \sim \mu}{\mathbb{E}} \underset{h^{\prime} \sim Q_{S'}^0}{\mathbb{E}}
f\left(S^{\prime},h^{\prime}\right)^{2}}_{:= \xi_m'}\right) \]

We first notice that $D_{2}\left(Q_{S} \| Q_S^0\right) = \sum_{i=1}^m D_2(\hat{Q}_{i+1}\| P_i)$ as our predictors are drawn independently once $S$ is given.

We also use that for any $i$, the Rényi divergence with $\alpha=2$ between $\hat{Q}_{i+1}$ and $P_i$ (two multivariate Gaussians with same covariance matrix) is $\frac{\|\hat{w}_{i+1}- w_i^0\|^2}{\sigma^2}$ (as recalled in \cite{gil2013renyi}).

We then remark that:

\[ \xi'_m = \underset{S^{\prime} \sim \mu}{\mathbb{E}} \underset{h^{\prime} \sim Q_{S'}^0}{\mathbb{E}}
\exp\left(2\lambda \left(\sum_{i=1}^m \mathbb{E}\left[\ell(h_i',z_i')\mid \mathcal{F}_{i-1}\right]- \sum_{i=1}^m \ell(h_i',z_i') \right)\right).  \]

Thus we recover the exponential moment $\xi_m$ from the Rivasplata's case up to a factor 2 within the exponential. We then apply \cref{l: exp_moment_online} with $\lambda'= 2\lambda$ to obtain that $\xi_m'\leq \exp\left(  2\lambda^2K^2m \right)$.

Combining all those facts allows us to conclude.

\subsubsection{Proof of \cref{cor: OPBD_test_bound}}

We apply the exact same proof than \cref{cor: OPBD_optim_funcs}. The only difference is the way to define our stochastic kernels. We now take, for a single online predictive sequence $\hat{Q}$ the following stochastic kernels:

We consider the stochastic kernel $Q$ belonging to the class $\mathcal{C}$ defined in \cref{eq: class_kernels} such that for any $S\in\mathcal{S}, Q(S,.) = \hat{Q}_1\otimes ... \otimes \hat{Q}_{m}$ and we take $Q_0=Q$.

This fact allows the divergence terms (Rényi or KL depending on which bound we consider) to vanish. The rest of the proof remains unchanged.

\section{Additional experiment}
\label{sec: error_bars}

In this section we perform error bars for our OPBD methods in order to evaluate their volatility.
We ran $n=50$ times our algorithms and then show in the table below for each data set the means and the standard deviation of our averaged cumulative losses at regular time steps. We denote for $i\in\{1,2\}$ 'OPBD $\Psi_i$' to indicate that this algorithm is our OPBD method used with thev optimisation objective $\Psi_i$.

\begin{table}[]
\label{tab: error_bars}
\begin{tabular}{|l|l|l|l|l|}
\hline
       & means OPBD $\Psi_1$ & std OPBD $\Psi_1$ & means OPBD $\Psi_2$ & std OPBD $\Psi_2$ \\ \hline
t=200  &      0.2014           &      0.0034         &    0.1993                 &     0.0007              \\ \hline
t=400 &      0.1888           &       0.0030        &    0.1861                 &       0.0004            \\ \hline
t=600  &     0.1867            &       0.0023        &       0.1839              &         0.0003          \\ \hline
t=800 &     0.1714            &         0.0020      &    0.1686                 &    0.0003               \\ \hline
t=1000 &    0.1760             &        0.0016       &            0.1731         &         0.0003          \\ \hline
\end{tabular}
\caption{Error bars for the Boston Housing dataset}

\begin{tabular}{|l|l|l|l|l|}
\hline
       & means OPBD $\Psi_1$ & std OPBD $\Psi_1$ & means OPBD $\Psi_2$ & std OPBD $\Psi_2$ \\ \hline
t=100  &      0.1619           &      0.0063         &    0.1601                 &     0.0030              \\ \hline
t=200 &      0.1350           &       0.0057        &    0.1361                 &       0.0008            \\ \hline
t=300  &     0.1214            &       0.0044        &       0.1241              &         0.0009          \\ \hline
t=400 &     0.1210            &         0.0043      &    0.1238                 &    0.0021               \\ \hline
t=500 &    0.1131             &        0.0037       &            0.1159         &         0.0015          \\ \hline
\end{tabular}
\caption{Error bars for the Breast Cancer dataset}

\begin{tabular}{|l|l|l|l|l|}
\hline
       & means OPBD $\Psi_1$ & std OPBD $\Psi_1$ & means OPBD $\Psi_2$ & std OPBD $\Psi_2$ \\ \hline
t=150  &      0.7102           &      0.0061         &    0.7069                 &     0.0007              \\ \hline
t=300 &      0.6455           &       0.0056        &    0.6422                 &       0.0007            \\ \hline
t=450  &     0.6134            &       0.0042        &       0.6103              &         0.0007          \\ \hline
t=600 &     0.5860           &         0.0035      &    0.5837                 &    0.0008               \\ \hline
t=750 &    0.5685             &        0.0031       &            0.5664         &         0.0008          \\ \hline
\end{tabular}
\caption{Error bars for the PIMA Indians dataset}

\begin{tabular}{|l|l|l|l|l|}
\hline
       & means OPBD $\Psi_1$ & std OPBD $\Psi_1$ & means OPBD $\Psi_2$ & std OPBD $\Psi_2$ \\ \hline
t=4000  &      0.9320           &      0.0572         &    0.8905                 &     0.0003              \\ \hline
t=8000 &      0.6325           &       0.0335        &    0.5947                 &       0.0003            \\ \hline
t=12000  &     0.5314            &       0.0254        &       0.4954              &         0.0002          \\ \hline
t=16000 &     0.4967           &         0.0299      &    0.4477                 &    0.0004               \\ \hline
t=20000 &    0.5273            &        0.1056       &            0.4355         &         0.0030          \\ \hline
\end{tabular}
\caption{Error bars for the California Housing dataset}

\end{table}

\paragraph{Analysis} Those tables shows the robustness of our OPBD methods to their intrinsic randomness: we always have a decreasing mean through time as well as an overall variance reduction. Note that for the most complicated problem (California Housing dataset), the variance is the highest.
More precisely, we notice that the standard deviation of OPBD with $\Psi_1$ is always greater than the one of OPBD with $\Psi_2$ which is not a surprise as $\Psi_1$ involves a disintegrated KL divergence while $\Psi_2$ is a proper Rényi divergence. Hence the additional volatility for $OPBD with \Psi_1$.

This fact is particurlaly noticeable on the California Housing dataset where both the means and variance of OPBD with $\Psi_1$ increase drastically between t=16000 and t=20000 while the increase is more attenuated for OPBD with $\Psi_2$. This fact is also visible on \cref{fig: exp_results}.

\end{document}